%% file: main.tex
\documentclass[10pt]{article}

\usepackage{graphicx} 
\usepackage[utf8]{inputenc}
\usepackage[round]{natbib}

\usepackage{amsmath}

\usepackage{amsfonts}       
\usepackage{amssymb}
\usepackage{mathtools}
\usepackage{wrapfig}
\usepackage{booktabs}
\usepackage{caption}
\usepackage[hidelinks]{hyperref}
\usepackage{color}
\usepackage{pythonhighlight}
\usepackage{makecell}
\newenvironment{DIFnomarkup}{}{}

\usepackage{pifont}

\usepackage{mynotation} 

\usepackage[preprint]{TMLR/tmlr}

\title{Repulsive Ensembles for Bayesian Inference in Physics-Informed Neural Networks}

\author{\name Philipp Pilar \email philipp.pilar@chalmers.se \\
      \addr Department of Computer Science and Engineering\\
      Chalmers University of Technology and University of Gothenburg
      \AND
      \name Markus Heinonen \email markus.o.heinonen@aalto.fi\\
      \addr  Department of Computer Science\\
      Aalto University
      \AND
      \name Niklas Wahlstr{\"o}m \email niklas.wahlstrom@it.uu.se\\
      \addr  Department of Information Technology\\
      Uppsala University}

\def\month{}  
\def\year{} 
\def\openreview{}

\begin{document}

\maketitle

\begin{abstract}
Physics-informed neural networks (PINNs) have proven an effective tool for solving differential equations, in particular when considering the inverse problem.
When inferring solutions and parameters of the differential equation from data, uncertainty estimates are preferable to point estimates, as they give an idea about the accuracy of the solution.
In this work, we consider the inverse problem for a selection of basic differential equations and employ repulsive ensembles of PINNs (RE-PINN) to obtain such estimates.
The repulsion is implemented by adding a particular repulsive term to the loss function, which has the property that the ensemble predictions correspond to the true Bayesian posterior in the limit of infinite ensemble members.
Where available, we compare the ensemble predictions to Monte Carlo baselines.
Whereas the standard ensemble tends to collapse to maximum-a-posteriori solutions, the repulsive ensemble produces significantly more accurate uncertainty estimates and exhibits higher sample diversity.
The experimental results demonstrate that RE-PINN can accurately infer the Bayesian posterior in low-dimensional problems, establishing repulsive ensembles as a promising framework for uncertainty quantification in PINNs.
\end{abstract}

\section{Introduction}

Physics-informed neural networks were first introduced by \citet{lagaris1998artificial}, further developed by \citet{raissi2019physics}, and have become a thriving area of research since then
\citep{cuomo2022scientific, cai2021fluid, lawal2022physics, cai2021heat}.
They constitute an alternate approach to solving differential equations (DEs):
a neural network is trained to learn the solution, where the DE is included as a soft constraint during network training.
In practical applications, it is of particular importance to obtain uncertainty estimates for the model predictions.
This is typically addressed in a Bayesian framework \citep{box2011bayesian} and has led to the development of Bayesian neural networks (BNNs) \citep{neal2012bayesian, jospin2022hands}, and thereafter Bayesian PINNs (B-PINNs) \citep{yang2021bpinn}.
Since exact inference is intractable, alternative methods such as Markov Chain Monte Carlo (MCMC) \citep{neal2011mcmc}, variational inference (VI) \citep{graves2011practical}, and dropout \citep{gal2016dropout} are often used.

Deep ensembles \citep{lakshminarayanan2017simple} have emerged as a promising alternative to these approaches for uncertainty estimation.
Multiple networks are trained in parallel, and, due to randomness in initialization and training, they can end up in different local optima.
Assuming standard training losses, these solutions correspond to maximum-a-posteriori (MAP) estimates of the posterior.
To encourage diversity between the ensemble members, adding a repulsive term to the loss function that prevents them from collapsing to the same solution has proven fruitful \citep{liu2016stein}.
\begin{wrapfigure}{r}{0.42\textwidth}
    \vspace{0cm}
  \centering
  \includegraphics[width=0.95\linewidth]{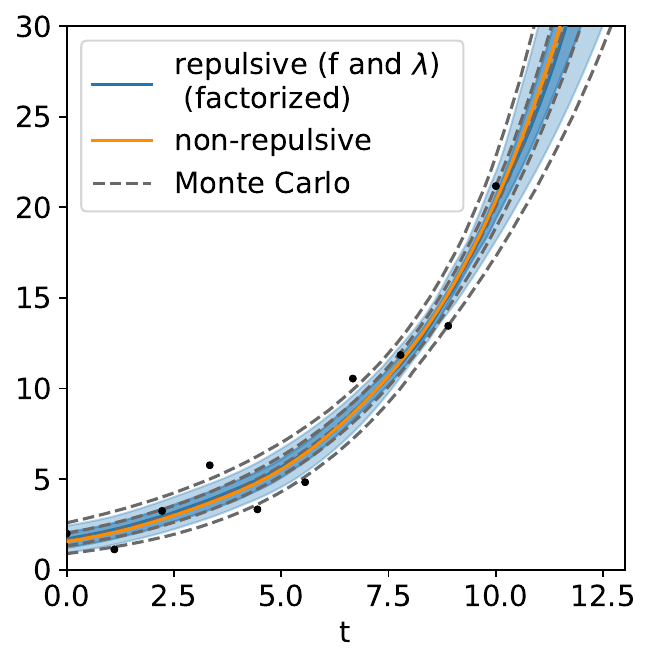}
    \caption{
    Distribution of the repulsive ensemble when solving the exponential equation.
    The ensemble median is depicted, together with the 0.1-0.9 quantiles, and the 0.25-0.75 quantiles (shaded areas).
    The non-repulsive ensemble collapses to the MAP estimate.
    The dashed lines correspond to the MC baseline.
    The black dots depict training data.
    The ensembles have 50 members.
    }
    \label{fig:ds1_front}  
    \vspace{0cm} 
\end{wrapfigure}
Due to the non-identifiability of neural networks \citep{roeder2021linear}, i.e., the fact that different configurations of the network parameters (consider, e.g., permutations of hidden units) can lead to the same predictions in function space, it is especially effective to add this repulsion directly in function space \citep{wang2019function}, rather than in weight space.
\citet{d2021repulsive} then gave justification for repulsive ensembles from the Bayesian point of view.
Building on connections between particle methods and deep ensembles \citep{jordan1998variational, wang2019function}, they derived a specific form for the repulsive term that guarantees convergence of the ensemble distribution to the true Bayesian posterior in the limit of infinite ensemble members.

In this work, we develop a method for uncertainty estimation in PINNs that corresponds to the Bayesian posterior.
That is, the method quantifies the uncertainty in the predictions that arise due to finite and noisy data.
We consider the inverse problem, where the PINN learns the DE solution jointly with the DE parameters from data.
Deep ensembles constitute a promising candidate for this task:
since each ensemble member is individually constrained to fulfill the DE with its own set of DE parameters, each will constitute a valid DE solution.
In turn, information about the DE solution space will be captured in the uncertainty intervals.
This would, in general, not be the case when using VI, which requires the posterior to belong to a predefined family of variational distributions, nor when using dropout techniques to estimate uncertainty, which have been shown to be detrimental to PINN performance \citep{yang2021bpinn}.
In this work, we apply the approach of \citet{d2021repulsive} to PINNs and investigate how its properties are affected by the additional DE loss term and parameters.
We show that repulsion needs to be applied in the joint space of function values and DE parameters, according to theory.
We conduct experiments on different low-dimensional problems and evaluate the effectiveness of different approximations of the repulsive term.

In Figure \ref{fig:ds1_front}, the performance of our approach is demonstrated for the exponential differential equation.
It is apparent that the repulsive ensemble of PINNs (RE-PINN) agrees very well with the Monte Carlo (MC) baseline, both in terms of the median prediction as well as the uncertainty intervals, whereas the non-repulsive ensemble essentially collapses to the MAP estimate.

\section{Related work}

\citet{yang2021bpinn} introduce a Bayesian formulation of PINNs (B-PINN) that builds on Bayesian NNs \citep{neal2012bayesian, jospin2022hands}, and where PDE residuals are treated as low-noise measurements.
They show that Hamiltonian Monte Carlo gives good uncertainty estimates, whereas variational inference (VI) with mean-field Gaussian approximation, and dropout lead to unreasonable estimates.
In \citet{li2024bayesian}, they apply B-PINN with VI to seismic data.

In \citet{jiang2023practical}, ensembles of PINNs (E-PINNs) are considered for solving inverse heat conduction problems.
They combine the PINN training with an adaptive sampling scheme to improve the predictive capabilities of the model.
In \citet{sahli2020physics}, ensembles of PINNs with randomized priors are used for active learning in cardiac activation mapping.
The uncertainty estimates obtained from the ensemble are utilized to suggest new measurements.

In \citet{haitsiukevich2023improved}, ensembles of PINNs are used as a means to improve the quality of the PINN solution.
The agreement between the ensemble members is used as a metric to ensure accurate solutions before expanding the domain of the solution.
In \citet{zou2025learning}, they propose ensembles of PINNs as a means to solve ODEs and PDEs with solution multiplicity, since ensemble members with different initializations can converge to different solutions.
In \citet{fang2023ensemble}, they develop a technique inspired by classical gradient-boosting techniques; 
the members of a PINN ensemble are trained sequentially to improve upon the previous predictions.

In \citet{yang2022multi}, confidence intervals for PINNs are obtained by having the PINN predict multiple (hundreds of) outputs, which are supposed to reflect the uncertainty in the prediction.
In \citet{tan2025evidential}, the principles of evidential deep learning are employed to obtain uncertainty estimates.

In \citet{rover2024pinnferring}, repulsive ensembles of PINNs are used to estimate the Hubble function from supernova data.
While they also employ repulsive ensembles to obtain uncertainty estimates, they do not infer DE parameters and only apply repulsion in function space.
The different works discussed in this section and the methods of uncertainty quantification considered in them are summarized in Table \ref{tab:papers}.

\begin{table}[t]
    \centering
    \resizebox{\textwidth}{!}{
    \begin{tabular}{l c c c c c r}
    \toprule
    Method & MC & VI & ensemble & repulsive ensemble & other & Reference \\
    \midrule
    B-PINN & \cmark & \cmark &  &    &   & \citet{yang2021bpinn} \\
    BPINN-VI &    & \cmark &  &    &  & \citet{li2024bayesian} \\
    MO-PINN & \gmark &  &  &  & \cmark & \citet{yang2022multi} \\
    E-PINN &  &  & \cmark &    &   & \citet{jiang2023practical} \\
    Multi-solution PINN &  &  & \cmark &    &   & \citet{zou2025learning} \\
    EikonalNet &  &  & \cmark &  &  & \citet{sahli2020physics} \\
    Functional repulsion PINN &  &  & \gmark & \cmark  &  & \citet{rover2024pinnferring} \\
    RE-PINN & \gmark & \gmark & \gmark & \cmark &  & our paper \\
    \bottomrule
    \end{tabular}
    }
    \caption{
    Overview of works that consider uncertainty quantification in PINNs.
    Green ticks indicate methods that constitute integral parts of the method described in the paper.
    Grey ticks indicate methods that were considered for comparison.
    }
    \label{tab:papers}
\end{table}

\section{Background: PINNs in the Bayesian Setting}

In this paper, we employ PINNs \citep{raissi2019physics} to solve differential equations of the form
\begin{equation} \label{eq:DE}
    \mathcal{F}(f, \lambda)(x) = 0,
\end{equation}
where $\mathcal{F}$ is the differential operator defining the differential equation with parameters $\lambda$, and where $f(x)$ is the solution of the differential equation at inputs $x$, which often contain time or spatial coordinates.
For example, the differential operator corresponding to the exponential equation from Figure \ref{fig:ds1_front} is given by $\mathcal{F}(f, \lambda)(x) = \frac{d f(x)}{d x} - \lambda f(x)$.
Both $f$ and $\lambda$ are to be inferred from data.
The solution is modeled as a neural network $\hat f(x) := \hat f(x; \theta)$ with network parameters $\theta$.
Model predictions for the DE parameters are denoted by $\hat \lambda$.

We follow the approach of \citet{yang2021bpinn} to describe PINNs in the Bayesian setting.
The data is given by $\mathcal{D}_d = \left\{(x_j, y_j) \right\}_{j=1}^{N_d}$, where the $y_j$ denote noisy measurements of $f(x_j)$.
In addition, we choose another dataset $\mathcal{D}_c = \left\{(x_j, 0)\right\}_{j=1}^{N_c}$, which contains collocation points at which fulfillment of the DE is enforced during training.
The zeros in this dataset are to be viewed as noisy measurements of the right-hand side of~\eqref{eq:DE} with standard deviation $\sigma_{\mathcal{F}}$.
This interpretation captures the soft-constraint nature of PINNs, which allows for slight deviations of the DE residuals from zero.
The two datasets are then collected in $\mathcal{D} = \left\{ \mathcal{D}_d, \mathcal{D}_c \right\}$.

Since we will eventually consider repulsion in function space, we directly consider the distribution of function values at discrete input points of interest, collected in the vector $\f$.
The posterior distribution $p(\f, \lambda|\mathcal{D})$ is employed as a means to systematically determine the loss function.
Bayes' law states that
\begin{equation} \label{eq:Bayes}
p(\f, \lambda|\mD) = \frac{p(\mD|\f, \lambda)p(\f, \lambda)}{p(\mD)},
\end{equation}
where $p(\mD|\f, \lambda)$ is the likelihood, $p(\f, \lambda)$ the prior distribution, and $p(\mD)$ the evidence.
After taking the negative logarithm, we obtain
\begin{align} \label{eq:L_BPINN}
    \mathcal{L} 
                &= \mathcal{L}_f + \mathcal{L}_{\mathcal{F}} - \log p(\f, \lambda) + \log p(\mD),
\end{align}
where $\mathcal{L}_f$ and $\mathcal{L}_{\mathcal{F}}$ denote the likelihoods for the different data modalities at hand.
Assuming Gaussian noise and omitting constants, they are given by

\noindent
\begin{subequations} \label{eq:PINN_losses}
\begin{minipage}{0.45\textwidth}
  \begin{align}  \label{eq:Lf} 
    \mathcal{L}_f &= \frac{1}{2\sigma_f^2} \sum_{j=1}^{N_d} (\hat f(x_j) - y_{j})^2,
  \end{align}
\end{minipage}
\hfill
\begin{minipage}{0.45\textwidth}
  \begin{align}  \label{eq:Ll} 
    \mathcal{L}_{\mathcal{F}} = \frac{1}{2\sigma_{\mathcal{F}}^2} \sum_{j=1}^{N_{c}}  (\mathcal{F}(\hat f, \hat \lambda)(x_j))^2.
  \end{align}
\end{minipage}
\end{subequations}

In case of different noise types, \eqref{eq:Lf} would need to be replaced with the corresponding negative log-likelihood.
The loss \eqref{eq:Ll} is equivalent to the soft loss term from the original PINN formulation \citep{raissi2019physics}.
If prior knowledge about the problem is available, the prior distribution entering the loss function can be chosen accordingly.
Otherwise, a uniform distribution can be chosen, resulting in a constant loss term.
The evidence term likewise remains constant.

\section{Method} \label{sec:method}

The aim of our method is to obtain an estimate of the Bayesian posterior \eqref{eq:Bayes} for the function values $f$ and DE parameters $\lambda$.
When training an ensemble of neural networks with loss function \eqref{eq:L_BPINN}, this will not yield samples from the Bayesian posterior, but instead a collection of MAP estimates \citep{lakshminarayanan2017simple}.

\citet{d2021repulsive} addressed this issue for standard neural networks.
By considering the Wasserstein gradient flow \citep{ambrosio2008gradient} when minimizing the KL-divergence between the ensemble distribution $\rho(\f)$, and the true posterior $p(\f|\mathcal{D})$, they derived an expression for the function value updates of the $i$-th ensemble member (see Appendix \ref{app:derivation} for details):
\begin{equation} \label{eq:particle_dynamics_f}
    \f_i^{t+1} = \f_i^{t} + \epsilon_t \left(\nabla_f \log p(\f_i^t|\mathcal{D}) - \nabla_f \log \rho(\f_i^t) \right).
\end{equation}
This expression now includes a term involving the density of the current ensemble predictions, which acts as a repulsive force between the ensemble members.
The vector $\f_i^t$ contains the function values $\hat f_i(x_j)$ of the $i$-th ensemble member at the data points $ x_j \in \mathcal{D}_d$.
This discretization is necessary since continuous functions are intractable.

When considering the inverse problem in PINNs, posterior estimates for $\lambda$ as well as $f$ are needed.
Due to the qualitative differences between the network parameters $\theta$, which result in the function values $\f$, and the DE parameters $\lambda$, which only show up in the loss \eqref{eq:Ll}, the modification of \eqref{eq:particle_dynamics_f} requires careful consideration.
The key point is that $\f$ and $\lambda$ do not explicitly depend on each other in PINNs.
Since the derivation of \eqref{eq:particle_dynamics_f} does not make restrictive assumptions on the quantities involved, this means that an analogous expression to \eqref{eq:particle_dynamics_f} can be obtained by substituting $[\f^\Transp, \lambda^\Transp]^\Transp$ for $\f$:
\begin{equation} \label{eq:update_discrete_0}
    \begin{bmatrix}
        \f_i^{t+1} \\ 
        \lambda_i^{t+1}
    \end{bmatrix}
    =    
    \begin{bmatrix}
        \f_i^{t} \\ 
        \lambda_i^{t}
    \end{bmatrix}
    + \epsilon_t
    \begin{bmatrix}
        \nabla_f ( \log p(\f_i^t, \lambda_i^t|\mathcal{D}) - \log \rho(\f_i^t, \lambda_i^t)\big)  \\
        \nabla_{\lambda} (\log p(\f_i^t, \lambda_i^t|\mathcal{D}) -  \log \rho(\f_i^t, \lambda_i^t)\big)
    \end{bmatrix}.
\end{equation}
Making use of \eqref{eq:Bayes} and plugging in the expressions from \eqref{eq:PINN_losses}, we obtain (after omitting constants)
\begin{equation}
    \begin{bmatrix}
        \f_i^{t+1} \\ 
        \lambda_i^{t+1}
    \end{bmatrix}
    =    
    \begin{bmatrix}
        \f_i^{t} \\ 
        \lambda_i^{t}
    \end{bmatrix}
    + \epsilon_t
    \begin{bmatrix}
        \nabla_f \big(-\mathcal{L}_{fi}  - \mathcal{L}_{\mathcal{F}i} + \log p(\f_i^t, \lambda_i^t) - \log \rho(\f_i^t, \lambda_i^t)\big)  \\
        \nabla_{\lambda} \big( -\mathcal{L}_{\mathcal{F}i}  + \log p(\f_i^t, \lambda_i^t) - \log \rho(\f_i^t, \lambda_i^t)\big) 
    \end{bmatrix}.
\end{equation}
Since neural networks are not trained directly in function space, an expression for the corresponding updates in weight space is required.
The gradient in $\theta$ can be obtained straightforwardly by utilizing the chain rule, resulting in:
\begin{equation} \label{eq:update_discrete}
    \begin{bmatrix}
        \theta_i^{t+1} \\ 
        \lambda_i^{t+1}
    \end{bmatrix}
    =    
    \begin{bmatrix}
        \theta_i^{t} \\ 
        \lambda_i^{t}
    \end{bmatrix}
    + \epsilon_t
    \begin{bmatrix}
        \left( \Jft{i}{t} \right)^\Transp \nabla_f \big(-\mathcal{L}_{fi}  - \mathcal{L}_{\mathcal{F}i} + \log p(\f_i^t, \lambda_i^t) - \log \rho(\f_i^t, \lambda_i^t)\big) \\
        \nabla_{\lambda} \big( -\mathcal{L}_{\mathcal{F}i}  + \log p(\f_i^t, \lambda_i^t) - \log \rho(\f_i^t, \lambda_i^t)\big) 
    \end{bmatrix},
\end{equation}
where $\Jft{i}{t}$ denotes the Jacobian.
Since neither $\lambda$ enters directly into $\f$, nor vice versa, the expression for the $\lambda$ updates remains unchanged, and no $\lambda$-derivatives are required when calculating the Jacobian of $\f$.
Both quantities interact only indirectly, through the likelihood term $\mathcal{L}_{\mathcal{F}}$, as well as the prior and the repulsive term.

When training our ensemble of PINNs, \eqref{eq:update_discrete} implies the following loss function for the $i$-th ensemble member:
\begin{equation}
    \mathcal{L}_i = \mathcal{L}_{fi}  + \mathcal{L}_{\mathcal{F}i} - \log p(\f_i, \lambda_i) + \log \rho (\f_i, \lambda_i).
\end{equation}
The Jacobian from \eqref{eq:update_discrete} does not show up explicitly in the loss, since it will automatically be taken into account through the chain rule.
After normalizing the loss with respect to the number of data points $N_d$ and rescaling, we obtain
\begin{align} \label{eq:loss}
    \tilde{\mathcal{L}}_i = \, &\frac{1}{N_d} \sum_{j=1}^{N_d}  \left(\hat f_i(x_j) - y_j \right)^2 + \frac{f_{\lambda}}{N_c} \sum_{j=1}^{N_c}  \left(\mathcal{F}( \hat f_i(x_j), \lambda_i)\right)^2 \\ 
    &+ f_{\rho} \big(-\log p(\f_i, \lambda_i) + \log \rho (\f_i, \lambda_i) \big), \nonumber
\end{align}
where $f_{\lambda} = \frac{\sigma_f^2 N_c}{\sigma_{\mathcal{F}}^2 N_d}$ and $f_{\rho} = \frac{2 \sigma_f^2}{N_d}$.

Evaluating the loss \eqref{eq:loss} requires knowledge of the density $\rho$, which is unavailable.
Hence, we utilize a particle approximation, that is, we approximate the distribution via the ensemble members.
For this, we have used kernel density approximation (KDE, see Appendix \ref{app:KDE}).
As pointed out in \citet{d2021repulsive}, KDE becomes exact in the limit of infinite ensemble members;
this implies that minimizing the loss \eqref{eq:loss} would lead to the true Bayesian posterior, also in our case.

\begin{table}[t]
\centering
\begin{tabular}{l|l}
   \hline
   Repulsion & repulsive term \\
   \hline
   non-repulsive &  $\log \rho(\f, \lambda) \rightarrow 0$  \\
   repulsive ($f$)  &  $\log \rho(\f, \lambda) \rightarrow \log \rho(\f)$  \\ 
   repulsive ($\lambda$)  &  $\log \rho(\f, \lambda) \rightarrow \log \rho(\lambda)$  \\ 
   repulsive ($f,\lambda$) &  $\log \rho(\f, \lambda)$ \\ 
   repulsive ($f,\lambda$) - factorized &  $\log \rho(\f, \lambda) \approx \log \rho_f(\f) + \log \rho_{\lambda}(\lambda)$  \\ 
   repulsive ($f,\lambda$) -  fully factorized & $\log \rho(\f, \lambda) \approx \frac{1}{\sqrt{N_d}} \left(  \sum_j \log \rho_{fj}(f_{j})  + \sum_l  \log \rho_{\lambda l}(\lambda_{l}) \right)$ \\
   \hline
\end{tabular}
\caption{
Different variants of repulsive ensembles that we considered in our experiments.
The column to the right gives the expression used for the repulsive term.
The first three models denote variants that were implemented for the sake of comparison, where repulsion is either omitted or only takes place in a subspace.
In the fourth model, the full joint density is kept in the repulsive term without approximation.
The last two models correspond to different ways of approximating the full repulsive term by factorizing the density.
}
\label{tab:repulsive_models}
\end{table}

The prior $p(\f, \lambda)$ can be chosen freely and in a way that the evaluation will be straightforward.
The approximation of the density $\rho(\f, \lambda)$ via KDE, however, needs special consideration.
Since $\f$ and $\lambda$ are rather different in nature, this may affect the performance of a naive approximation via KDE.
One way of improving the approximation could be to approximate the density as $\rho(\f, \lambda) \approx \rho_f(\f) \rho_{\lambda}(\lambda)$;
then, KDE could be performed separately for $\f$ and $\lambda$.
Furthermore, the dimensionality of $\f$ may be large, which can also lead to bad KDE estimates.
Factorizing the density further between different function values could improve the estimates.

In Table \ref{tab:repulsive_models}, we summarize the various approximations and modifications for the density term that we consider in our experiments.
We also include variants where the repulsion happens in only one of the two spaces, to clearly isolate the impact that repulsion in $f$-space and $\lambda$-space has, respectively.
The fully-factorized variant has the advantage that only one-dimensional distributions need to be approximated with KDE.
However, this also means that correlations between function values $f$ at different points get lost, which will typically result in too strong repulsion.
To counteract this issue, it has proven effective to normalize the repulsive term with the factor $\frac{1}{\sqrt{N_d}}$ for this model in our experiments.
This factor is heuristically motivated in analogy with the factor that appears in the elementary formula for the standard deviation.
In DE solutions, uncertainty at a given point is also influenced by all available measurements.
Empirically, we have found this scaling factor to produce excellent results.

\section{Experiments} \label{sec:experiments}

\begin{DIFnomarkup}
\begin{table} [t]
\resizebox{\textwidth}{!}{%
    \centering
    \includegraphics[width=1\linewidth]{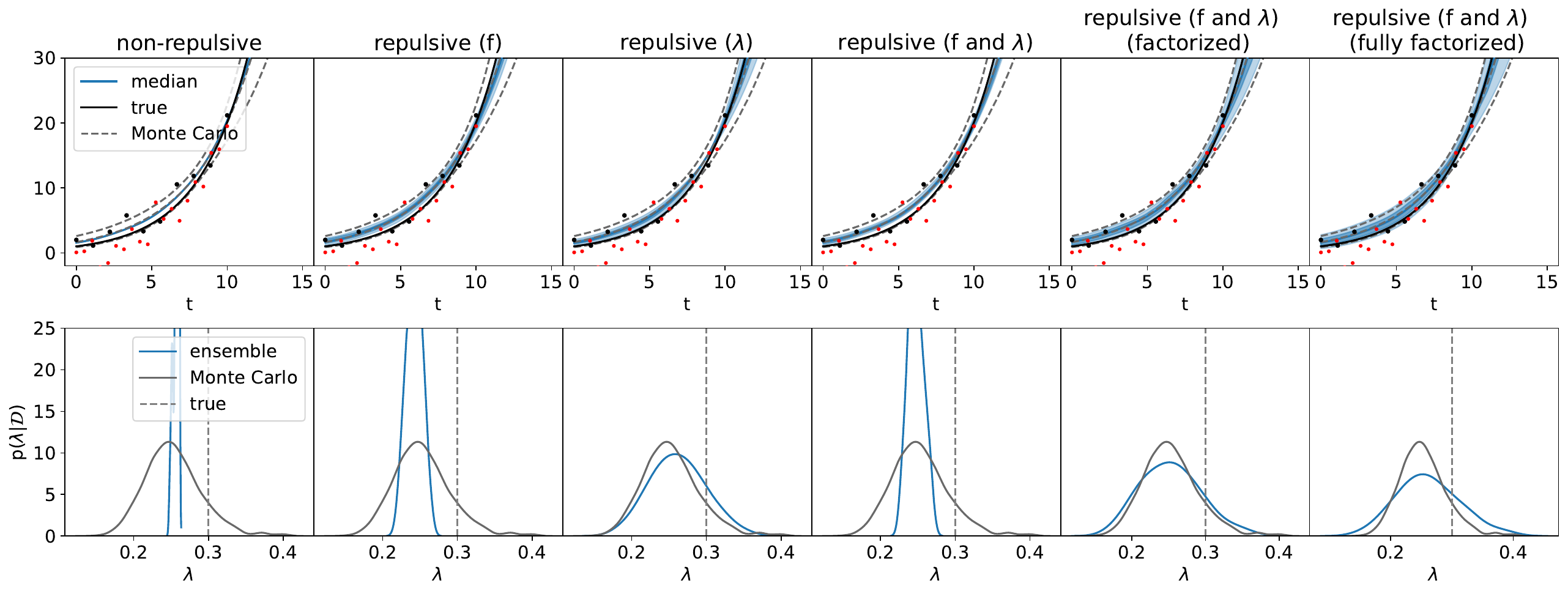}}
    \captionof{figure}{ 
    Solving the exponential differential equation \eqref{eq:exp} with repulsion in ($f$, $\lambda$)-space.
    \textbf{Top row:}
    The predictions in $f$-space are depicted.
    The light- and dark-shaded areas give the regions between the [0.1, 0.9]- and the [0.25, 0.75]-quantiles, respectively.
    The dashed lines depict the 0.1- and 0.9- quantiles as obtained via MCMC, together with the median.
    Black dots correspond to train data and red dots to test data.
    \textbf{Bottom row:}
    The predictions in $\lambda$-space are depicted.
    The ensemble distributions are compared to the distribution obtained via MCMC.
    The ensembles have 50 members.
    }
    \label{fig:ds1_MC_comp}
\resizebox{\textwidth}{!}{%
\begin{tabular}{l c c c c c c }
\toprule
Repulsion & RMSE (true) & logL/N (test) & $|\lambda - \hat \lambda|$ & logL $\lambda$ & W (f) & W $(\lambda)$ \\ 
\midrule 
non-repulsive & \phantom{}0.62$\pm$0.27\phantom{}\phantom{} & \phantom{}-60564.5$\pm$93742.0\phantom{-}\phantom{} & \phantom{}0.029$\pm$0.011\phantom{}\phantom{} & \phantom{}-541.58$\pm$969.23\phantom{-}\phantom{} & \phantom{}2.5$\pm$0.2\phantom{}\phantom{} & \phantom{}0.030$\pm$0.001\phantom{}\phantom{}\\
repulsive ($f$) & \phantom{}0.69$\pm$0.27\phantom{}\phantom{} & \phantom{}-26.0$\pm$8.2\phantom{-}\phantom{0} & \phantom{}0.034$\pm$0.021\phantom{}\phantom{} & \phantom{}-10.82$\pm$11.51\phantom{-}\phantom{} & \phantom{}2.1$\pm$0.4\phantom{}\phantom{} & \phantom{}0.028$\pm$0.004\phantom{}\phantom{}\\
repulsive ($\lambda$) & \textbf{\phantom{}0.59}$\pm$0.26\phantom{}\phantom{} & \phantom{}-64.7$\pm$21.9\phantom{-}\phantom{} & \textbf{\phantom{}0.024}$\pm$0.010\phantom{}\phantom{} & \textbf{\phantom{}1.98}$\pm$0.13\phantom{}\phantom{} & \phantom{}0.9$\pm$0.1\phantom{}\phantom{} & \textbf{\phantom{}0.007}$\pm$0.002\phantom{}\phantom{}\\
repulsive ($f,\lambda$) & \phantom{}0.66$\pm$0.28\phantom{}\phantom{} & \phantom{}-48.0$\pm$18.7\phantom{-}\phantom{} & \phantom{}0.032$\pm$0.018\phantom{}\phantom{} & \phantom{}-3.69$\pm$7.69\phantom{-}\phantom{} & \phantom{}2.1$\pm$0.3\phantom{}\phantom{} & \phantom{}0.026$\pm$0.003\phantom{}\phantom{}\\
repulsive ($f,\lambda$) (factorized) & \phantom{}0.65$\pm$0.26\phantom{}\phantom{} & \phantom{}-11.7$\pm$3.9\phantom{-}\phantom{0} & \phantom{}0.027$\pm$0.017\phantom{}\phantom{} & \phantom{}1.78$\pm$0.17\phantom{}\phantom{} & \textbf{\phantom{}0.8}$\pm$0.4\phantom{}\phantom{} & \phantom{}0.009$\pm$0.004\phantom{}\phantom{}\\
repulsive ($f,\lambda$) (fully factorized) & \phantom{}0.61$\pm$0.26\phantom{}\phantom{} & \textbf{\phantom{}-4.1}$\pm$0.8\phantom{-}\phantom{} & \phantom{}0.025$\pm$0.011\phantom{}\phantom{} & \phantom{}1.79$\pm$0.14\phantom{}\phantom{} & \phantom{}0.8$\pm$0.1\phantom{}\phantom{} & \phantom{}0.009$\pm$0.003\phantom{}\phantom{}\\
variational inference (B-PINN) & \phantom{}1.06$\pm$0.30\phantom{}\phantom{} & \phantom{}-64.7$\pm$36.5\phantom{-}\phantom{} & \phantom{}0.066$\pm$0.013\phantom{}\phantom{} & \phantom{}-7443.58$\pm$2528.78\phantom{-}\phantom{} & \phantom{}3.9$\pm$1.1\phantom{}\phantom{} & \phantom{}0.055$\pm$0.014\phantom{}\phantom{}\\
\bottomrule
\end{tabular}}
\captionof{table}{
Evaluation metrics when solving the exponential equation \eqref{eq:exp} with repulsion in ($f, \lambda$)-space, averaged over 5 runs with 50 ensemble members.
The mean values plus-or-minus one standard deviation are given.
The results for one of these runs are depicted in Figure \ref{fig:ds1_MC_comp}.
\\
}
\label{tab:ds1_Nx10_repf}
\end{table}
\end{DIFnomarkup}

In this section, we present the results obtained from different experiments to evaluate the performance of the approach.
The models were trained on noisy data,
\begin{equation}
    y(x) = f(x) + \epsilon,
\end{equation}
where $\epsilon \sim \mathcal{N}(0, \sigma_f^2)$.
Ensemble predictions are made by taking the ensemble mean, $\hat f(x) =  \frac{1}{N_e} \sum_{i=1}^{N_e} \hat f_i(x)$, and $\hat \lambda = \frac{1}{N_e} \sum_{i=1}^{N_e} \hat \lambda_i$, where $N_e$ is the number of ensemble members.
Details about the network architecture and training procedure are provided in Appendix \ref{app:training}.
The code for the project is available on GitHub \protect\footnote{\href{https://github.com/ppilar/RE-PINN}{https://github.com/ppilar/RE-PINN}}.


To evaluate the accuracy of the models, both in terms of the accuracy of the ensemble predictions, as well as the uncertainties, we employ a variety of evaluation metrics.
The root mean square error between the ensemble prediction and the true curve (RMSE (true)) quantifies the accuracy of the model.
To assess the quality of the ensemble uncertainties, we calculate the log-likelihood per data point on the test data (logL/N (test)).
The likelihood is calculated from a KDE approximation of the ensemble distributions via the ensemble members.
For the DE parameters $\lambda$, we calculate the absolute error $\lvert \lambda - \hat \lambda\rvert$ with respect to the true value, and logL $\lambda$, the log-likelihood of the true value given the ensemble distribution.
Furthermore, we compare how the B-PINN approach using variational inference (VI) from \citet{yang2021bpinn} performs on the same evaluation metrics.
An in-depth discussion of VI is given in Appendix \ref{app:VI}.

For the ODE examples, the fact that the solutions are uniquely determined by the initial conditions and ODE parameters provides the opportunity to perform Markov chain Monte Carlo (MCMC) in the corresponding, low-dimensional space.
In this way, a highly accurate proxy for the ground truth of the PINN posterior can be obtained.
We then calculated the Wasserstein distance between the MCMC and the ensemble distributions, both, $W (f)$ in function space (pointwise), and $W(\lambda)$ in $\lambda$-space.
Details on the MC algorithm are given in Appendix \ref{app:MC}.

Among these metrics, logL/N (test) is the most relevant in practice, as it can always be computed.
Higher values indicate that the model better explains unseen data and therefore correspond to better calibrated uncertainty estimates.
The remaining metrics are included to evaluate the absolute accuracy of the inferred posterior in our benchmark problems.
However, because they require either the true DE solution or a reference posterior obtained using a potentially more expensive method, they cannot, in general, be used in real-world applications.

\subsection{Exponential equation} \label{sec:exp}

\begin{table}
\resizebox{\textwidth}{!}{%
    \centering
    \includegraphics[width=1\linewidth]{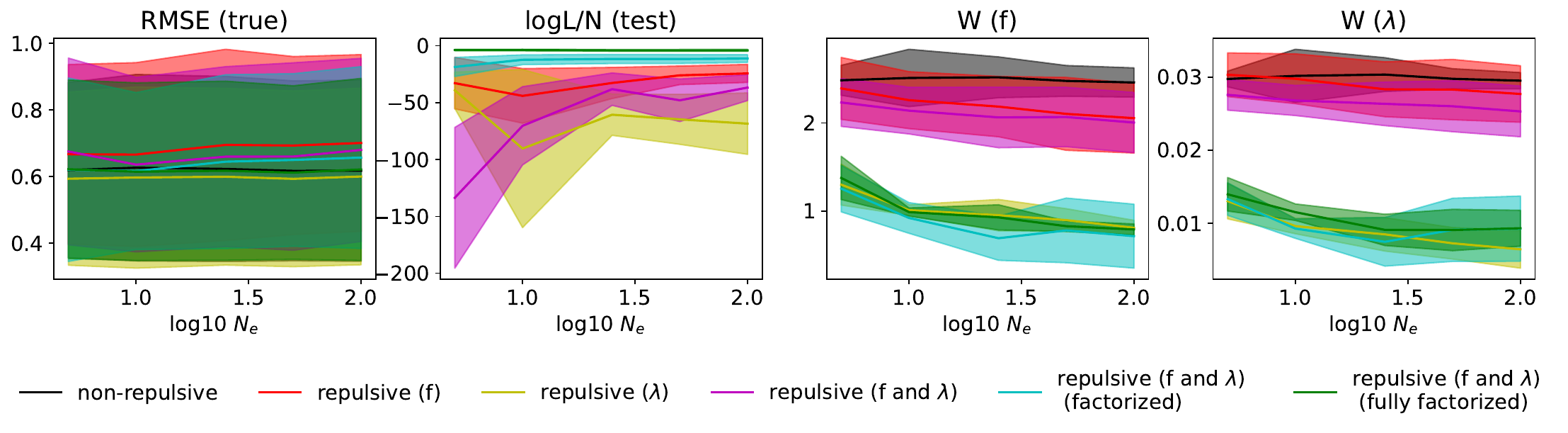}}
    \captionof{figure}{
    The evaluation metrics when solving the exponential equation with repulsion in ($f$, $\lambda$)-space with different numbers of ensemble members.
    For better visibility, the non-repulsive ensemble has been omitted for logL/N (test), since the values are far below those of the repulsive ensembles.
    }
    \label{fig:ds1_Ncomp_eval}
\end{table}

To illustrate the approach, we consider the example of the exponential differential equation, 
\begin{equation} \label{eq:exp}
    \frac{d f(t)}{d t} = \lambda f(t),
\end{equation}
which has the solution $f(t) = f_0 e^{\lambda t}$.
We solve the inverse problem, where $\lambda$ is determined jointly with the solution $f(t)$.
The factor $f_0$ is obtained automatically when evaluating $f(0)$.

In Figure \ref{fig:ds1_MC_comp}, the results for the different models are depicted, both in terms of the ensemble distribution in $f$-space, and in $\lambda$-space.
For the function space, the median of the ensemble is shown together with the 0.1, 0.25, 0.75, and 0.9 quantiles.
The corresponding 0.1, 0.9 quantiles and the median as obtained via MCMC are also depicted.
The corresponding evaluation metrics are given in Table \ref{tab:ds1_Nx10_repf}.
Plots of the individual ensemble member predictions can be found in Appendix \ref{app:exp}.

It is apparent that the standard, non-repulsive ensemble vastly underestimates the uncertainty in the prediction.
The ensemble members almost collapse to a point estimate.
In the plot showing the $\lambda$ distribution, we observe that the predictions are very close to the maximum of the Monte Carlo posterior.
This is expected, since standard neural network (and PINN) training converges to MAP solutions.
KDE in the high-dimensional space of function values does not seem to be very precise, as indicated by the results for repulsive ($f$) and repulsive ($f$ and $\lambda$), which still underestimate the uncertainties significantly.
The models with separate repulsion in $\lambda$-space perform notably better.
The factorized model in particular adheres well to the MCMC posterior, both in $f$- and $\lambda$-space.
The fully-factorized model still agrees well in $f$-space, but tends to overestimate the uncertainty in $\lambda$-space.

In terms of the evaluation metrics, the non-repulsive ensemble performs worst on all metrics apart from RMSE (true) and $|\lambda - \hat \lambda|$, which do not measure the uncertainty in the predictions but the accuracy of the mean.
The repulsive ensembles are on par with the non-repulsive ensemble when it comes to accuracy, and improve in terms of the uncertainty-quantifying metrics.
Whereas the repulsive ($\lambda$) model performs best on the metrics involving $\lambda$, the factorized and fully-factorized models perform better in function space.
The fully-factorized model achieves the highest likelihood on the test data.
VI performs clearly worse than the repulsive ensembles.

In Figure \ref{fig:ds1_Ncomp_eval}, evaluation metrics are given for the different models as a function of the number of ensemble members.
For metrics that quantify uncertainty, there is a clear tendency for the different variants of repulsive ensembles to improve with the number of ensemble members.
This is consistent with the fact that the approximation of the density $\rho$ via KDE becomes more accurate as the number of samples increases.
In the limit of infinite ensemble members, the repulsive ($f$ and $\lambda$) model is expected to give the true posterior, i.e., $W(f)$ and $W(\lambda)$ should go to values close to zero.
For the RMSE, on the other hand, the curves remain essentially flat.
This also makes sense, since the RMSE quantifies the accuracy of the mean prediction instead of the uncertainty.

\subsection{The damped harmonic oscillator} \label{sec:dHO}

\begin{figure}[t]
    \centering
    \includegraphics[width=1\linewidth]{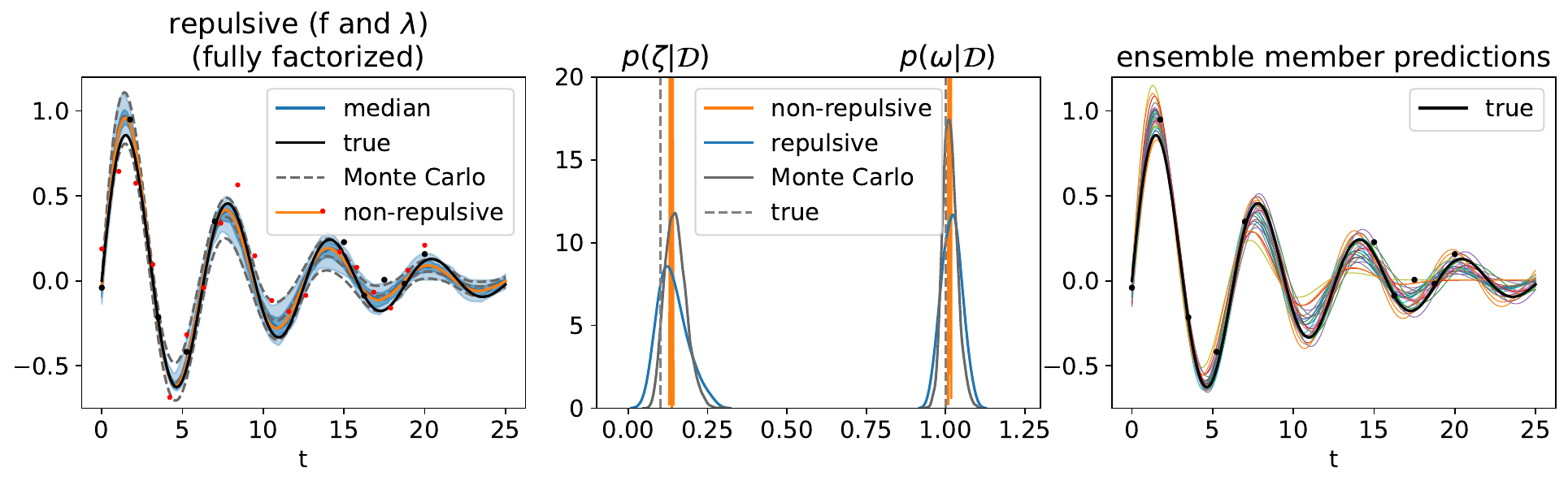}
    \captionof{figure}{ 
    Solving the damped harmonic oscillator \eqref{eq:dHO} with repulsion in ($f$, $\lambda$)-space, using the repulsive ensemble with fully-factorized density.
    \textbf{Left:}
    The predictions in $f$-space are depicted.
    The light- and dark-shaded areas give the regions between the [0.1, 0.9]- and the [0.25, 0.75]-quantiles, respectively.
    The dashed lines depict the 0.1- and 0.9- quantiles as obtained via MCMC, together with the median.
    The non-repulsive ensemble effectively collapses to the MAP estimate.
    Black dots correspond to train data and red dots to test data.
    \textbf{Middle:}
    The predictions in $\lambda$-space are depicted.
    The ensemble distribution is compared to the distribution obtained via MCMC.
    \textbf{Right:}
    The individual ensemble members are depicted.
    The ensembles have 25 members.
    }
    \label{fig:ds5_MC_comp}
\end{figure}
\begin{table} [!t]
\resizebox{\textwidth}{!}{%
\begin{tabular}{l c c c c c c }
\toprule
Repulsion & RMSE (true) & logL/N (test) & $|\lambda - \hat \lambda|$ & logL $\lambda$ & W (f) & W $(\lambda)$ \\ 
\midrule 
non-repulsive & \textbf{\phantom{}0.059}$\pm$0.013\phantom{}\phantom{} & \phantom{}-1498.9$\pm$822.7\phantom{-}\phantom{0} & \textbf{\phantom{}0.016}$\pm$0.004\phantom{}\phantom{} & \phantom{}-2903.7$\pm$2016.4\phantom{-}\phantom{} & \phantom{}0.045$\pm$0.004\phantom{}\phantom{} & \phantom{}0.024$\pm$0.006\phantom{}\phantom{}\\
repulsive ($f$) & \phantom{}0.059$\pm$0.014\phantom{}\phantom{} & \phantom{}-287.8$\pm$234.2\phantom{-}\phantom{} & \phantom{}0.016$\pm$0.004\phantom{}\phantom{} & \phantom{}-65.5$\pm$62.8\phantom{-}\phantom{} & \phantom{}0.041$\pm$0.005\phantom{}\phantom{} & \phantom{}0.022$\pm$0.006\phantom{}\phantom{}\\
repulsive ($\lambda$) & \phantom{}0.062$\pm$0.017\phantom{}\phantom{} & \phantom{}-17.3$\pm$3.3\phantom{-}\phantom{0} & \phantom{}0.021$\pm$0.007\phantom{}\phantom{} & \textbf{\phantom{}4.7}$\pm$0.4\phantom{}\phantom{} & \phantom{}0.027$\pm$0.004\phantom{}\phantom{} & \textbf{\phantom{}0.010}$\pm$0.004\phantom{}\phantom{}\\
repulsive ($f,\lambda$) & \phantom{}0.060$\pm$0.014\phantom{}\phantom{} & \phantom{}-860.5$\pm$841.0\phantom{-}\phantom{} & \phantom{}0.017$\pm$0.004\phantom{}\phantom{} & \phantom{}-263.8$\pm$247.6\phantom{-}\phantom{} & \phantom{}0.043$\pm$0.004\phantom{}\phantom{} & \phantom{}0.023$\pm$0.006\phantom{}\phantom{}\\
repulsive ($f,\lambda$) (factorized) & \phantom{}0.063$\pm$0.017\phantom{}\phantom{} & \phantom{}-13.4$\pm$1.1\phantom{-}\phantom{0} & \phantom{}0.021$\pm$0.007\phantom{}\phantom{} & \phantom{}4.7$\pm$0.4\phantom{}\phantom{} & \phantom{}0.026$\pm$0.004\phantom{}\phantom{} & \phantom{}0.010$\pm$0.004\phantom{}\phantom{}\\
repulsive ($f,\lambda$) (fully factorized) & \phantom{}0.061$\pm$0.016\phantom{}\phantom{} & \textbf{\phantom{}-2.0}$\pm$1.5\phantom{-}\phantom{} & \phantom{}0.020$\pm$0.006\phantom{}\phantom{} & \phantom{}4.6$\pm$0.3\phantom{}\phantom{} & \textbf{\phantom{}0.025}$\pm$0.005\phantom{}\phantom{} & \phantom{}0.013$\pm$0.003\phantom{}\phantom{}\\
variational inference (B-PINN) & \phantom{}0.131$\pm$0.035\phantom{}\phantom{} & \phantom{}-116.2$\pm$82.3\phantom{-}\phantom{0} & \phantom{}0.075$\pm$0.024\phantom{}\phantom{} & \phantom{}-239285.0$\pm$158340.7\phantom{-}\phantom{} & \phantom{}0.083$\pm$0.022\phantom{}\phantom{} & \phantom{}0.087$\pm$0.034\phantom{}\phantom{}\\
\bottomrule
\end{tabular}}
\captionof{table}{
Evaluation metrics when solving the damped harmonic oscillator \eqref{eq:dHO} with repulsion in ($f, \lambda$)-space, averaged over 5 runs with 25 ensemble members.
The mean values plus-or-minus one standard deviation are given.
The results for one of these runs are depicted in Figure \ref{fig:ds5_MC_comp}.
\\
}
\label{tab:dHO}
\end{table}

The DE for the damped harmonic oscillator is given by
\begin{equation} \label{eq:dHO}
    \frac{d f(t)^2}{d t^2} + 2 \zeta \omega \frac{d f(t)}{d t} + \omega^2 f(t) = 0,
\end{equation}
where $\omega$ denotes the undamped angular frequency and $\zeta$ the damping ratio.
In the underdamped regime, where $\zeta<1$, the solution is $f(t) = f_0 e^{-\zeta \omega t} \sin(\omega \sqrt{1-\zeta^2} t)$.

The results are given in Figure \ref{fig:ds5_MC_comp} and Table \ref{tab:dHO}.
A Figure depicting the ensemble distributions for all the different models is given in Appendix \ref{app:dHO}.
Again, the non-repulsive ensemble basically collapses to a single solution.
The repulsive model, on the other hand, approximates the MC baseline well overall.
Note that there is a gap in the data in the area $t \in [7, 15]$;
instead of a big blob that unconstrained methods might yield as the uncertainty estimate in this area, the uncertainty intervals, as obtained via the ensemble, retain information about the solution space of the DE, since it is inherent in the ensemble members.

When considering the posteriors for the DE parameters, we again find that the non-repulsive ensemble assigns exceedingly high certainty to values that are close to, but not exactly at, the true value.
The repulsive ensemble posterior again agrees reasonably well with the MC posterior.
In the rightmost plot, the individual ensemble members are shown.
Each of them constitutes a PINN solution in its own right.
While close to real solutions of the ODE, there can be small deviations due to the soft-constraint nature of the PINN.

The evaluation metrics in Table \ref{tab:dHO} show a very similar pattern to those in the previous example.
On the whole, the fully-factorized model performs best.

\subsection{The advection equation} \label{sec:adv}

\begin{wrapfigure}{r}{0.25\textwidth}
    \vspace{-1.3cm}
  \centering
  \includegraphics[width=\linewidth]{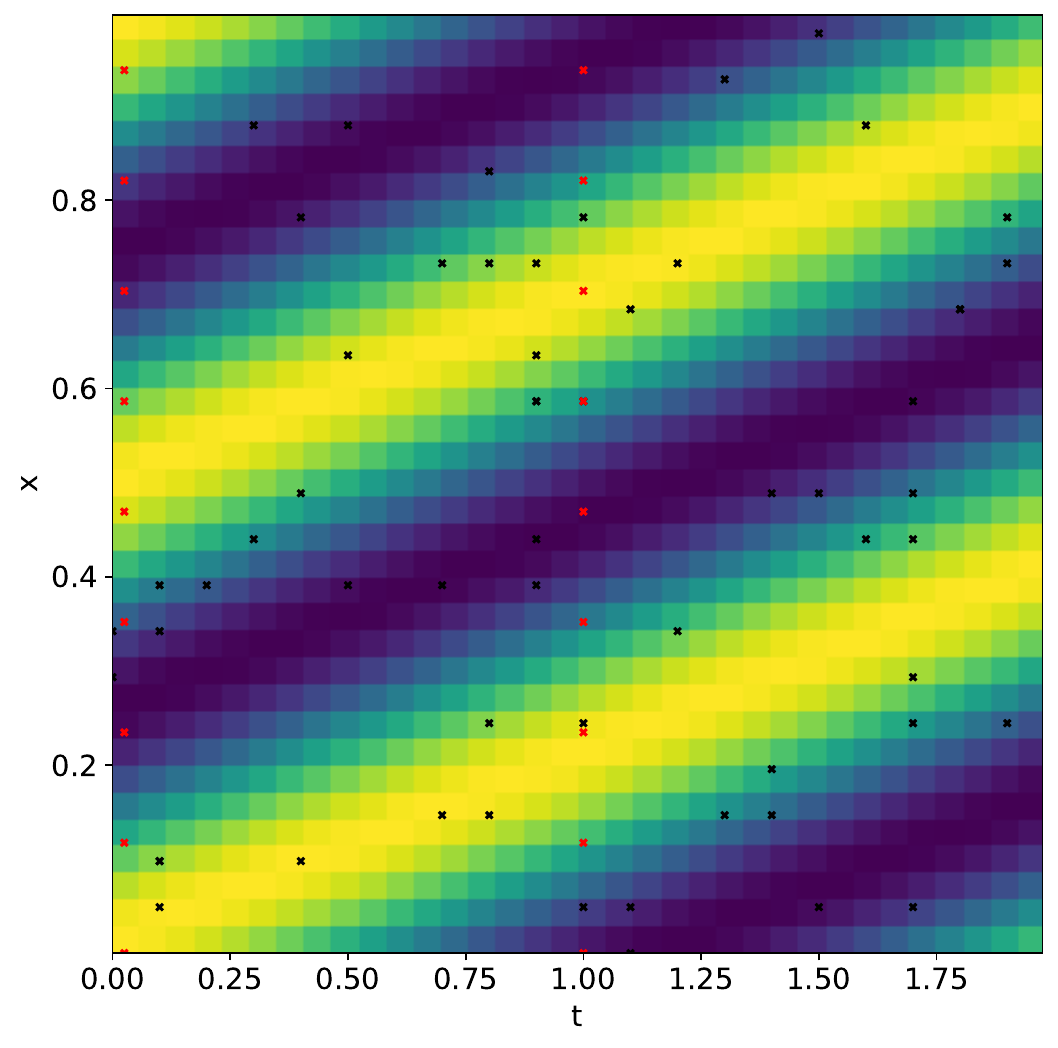}
    \caption{
    Example solution of the advection equation.
    Black dots correspond to training data and red dots to test data.
    }
    \label{fig:ds102_data}
    \vspace{-1.3cm}
\end{wrapfigure}
The advection equation is given by
\begin{equation} \label{eq:adv}
    \frac{\partial u(t,x)}{\partial t} + \lambda \frac{\partial u(t,x)}{\partial x} = 0.
\end{equation}
The solution for this PDE is given by $u(t,x)=u_0 (x - \lambda t)$, where $u_0$ denotes the initial conditions.

An illustration of the datasets under consideration is given in Figure \ref{fig:ds102_data}.
The data was obtained from \cite{takamoto2022pdebench}.
The results are given in Figure \ref{fig:ds102_comp} and Table \ref{tab:ds102_repf}.
The solutions in function space are illustrated along two cuts of the 2D domain. 
In function space, the difference between non-repulsive and repulsive ensembles is smaller than in the previous examples, but the non-repulsive ensemble still tends to underestimate uncertainty, especially in regions without data (see the right part of the $t=0.0$ curve).

For this example, no MC solution is available, due to the continuous boundary condition.
Nevertheless, from Table \ref{tab:ds102_repf}, we see that the repulsive ensembles perform significantly better on the log-likelihoods than the non-repulsive ensemble.
Altogether, the fully-factorized model again performs best.

\subsection{Comparison with variational inference} \label{sec:VI_comparison}

In this section, we evaluate how VI performs as a method for Bayesian inference in PINNs, using the B-PINN from \citet{yang2021bpinn}.
An in-depth discussion of the method is given in Appendix \ref{app:VI}.

In Figure \ref{fig:ds5_MC_comp_VI}, the performance of VI applied to the damped harmonic oscillator is illustrated.
The mean of the distribution follows the MC median reasonably well.
When it comes to sample diversity, VI performs clearly worse than the RE-PINN (compare Figure \ref{fig:ds5_MC_comp}).
The posterior distributions for the DE parameters are strongly squeezed to specific values, presumably because the mean-field Gaussian posterior does not allow for correlations between $\lambda$ and $\theta$, and this is the only way for the model to attain reasonable DE fulfillment.
The individual samples do not adhere to the DE consistently either.
For example, the maximum amplitude curve should be the same at each extremum, but this is not the case.
Instead, they mostly appear to be slightly shifted versions of the mean prediction.

From the results in Tables \ref{tab:ds1_Nx10_repf}-\ref{tab:ds102_repf}, it is apparent that VI performs worse in terms of uncertainty quantification than the RE-PINN.
While it does improve over the non-repulsive ensemble in terms of the log-likelihood on the test data, the log-likelihood on the DE parameters is significantly worse, indicating strongly squeezed posteriors.
Furthermore, the improvement in uncertainty quantification over the non-repulsive ensemble often comes at the cost of a degradation in the mean predictions as measured by RMSE (true) and $|\lambda - \hat \lambda|$.
A degradation in the sample quality is also reflected in higher mean values of the DE residuals, which are given in Table \ref{tab:de_residuals} in the appendix.
The likely reason for this is that the variational posterior does not include correlations between model parameters.
While better-suited posteriors could likely be designed, this would come at the cost of significant amounts of additional model tuning.

\begin{DIFnomarkup}
\begin{figure}
    \centering
    \includegraphics[width=\linewidth]{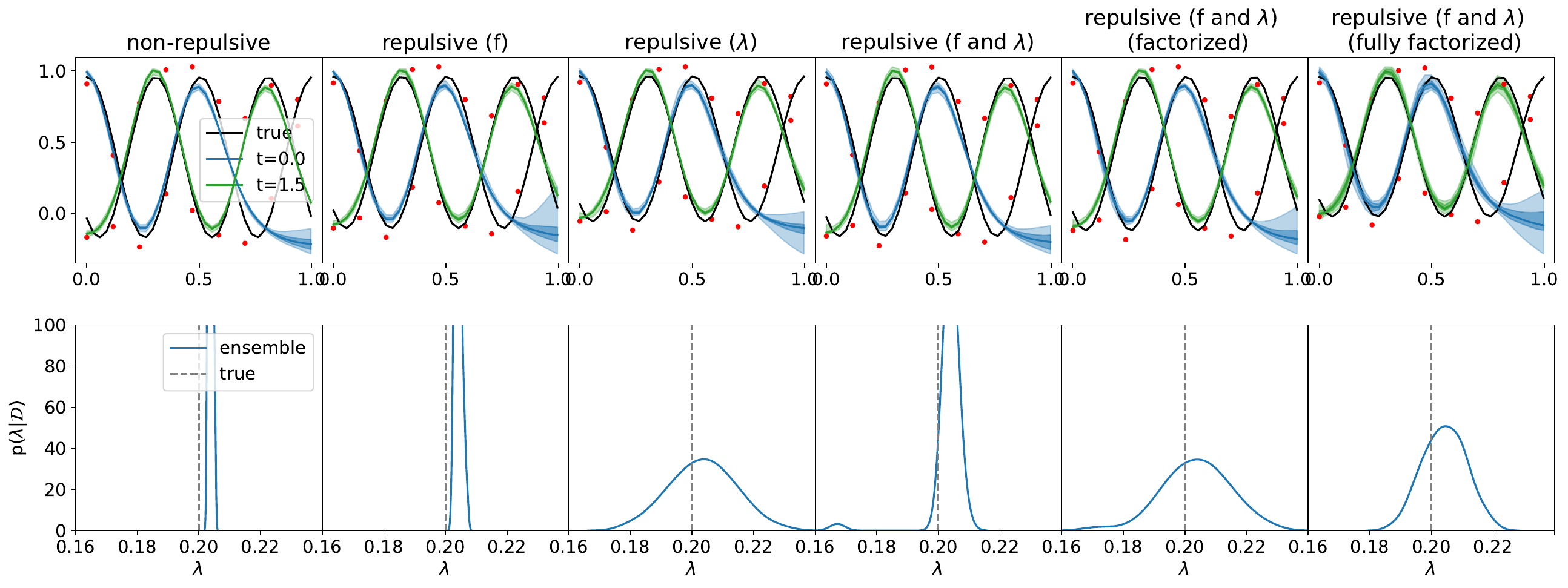}
    \caption{
    Solving the advection equation from \eqref{eq:adv} with repulsion in $(f,\lambda)$-space.
    The ensembles have 25 members.
    Red dots correspond to test data.}
    \label{fig:ds102_comp}
\end{figure}

\begin{table} [!t]
\centering
\resizebox{0.85\textwidth}{!}{%
\begin{tabular}{l c c c c }
\toprule
Repulsion & RMSE (true) & logL/N (test) & $|\lambda - \hat \lambda|$ & logL $\lambda$ \\ 
\midrule 
non-repulsive & \phantom{}0.125$\pm$0.062\phantom{}\phantom{} & \phantom{}-85.8$\pm$79.7\phantom{-}\phantom{} & \phantom{}0.020$\pm$0.030\phantom{}\phantom{} & \phantom{}-47.4$\pm$84.6\phantom{-}\phantom{}\\
repulsive (f) & \phantom{}0.122$\pm$0.058\phantom{}\phantom{} & \phantom{}-36.3$\pm$31.1\phantom{-}\phantom{} & \phantom{}0.020$\pm$0.030\phantom{}\phantom{} & \phantom{}-9.9$\pm$22.4\phantom{-}\phantom{}\\
repulsive ($\lambda$) & \phantom{}0.126$\pm$0.064\phantom{}\phantom{} & \phantom{}-44.0$\pm$33.1\phantom{-}\phantom{} & \phantom{}0.020$\pm$0.031\phantom{}\phantom{} & \phantom{}3.5$\pm$0.2\phantom{}\phantom{}\\
repulsive (f and $\lambda$) & \phantom{}0.123$\pm$0.061\phantom{}\phantom{} & \phantom{}-33.6$\pm$15.7\phantom{-}\phantom{} & \phantom{}0.018$\pm$0.029\phantom{}\phantom{} & \phantom{}-22.3$\pm$45.1\phantom{-}\phantom{}\\
repulsive (f and $\lambda$) (factorized) & \phantom{}0.121$\pm$0.058\phantom{}\phantom{} & \phantom{}-40.3$\pm$32.1\phantom{-}\phantom{} & \phantom{}0.020$\pm$0.031\phantom{}\phantom{} & \phantom{}3.5$\pm$0.2\phantom{}\phantom{}\\
repulsive (f and $\lambda$) (fully factorized) & \phantom{}0.121$\pm$0.057\phantom{}\phantom{} & \textbf{\phantom{}-3.3}$\pm$3.1\phantom{-}\phantom{} & \phantom{}0.020$\pm$0.031\phantom{}\phantom{} & \textbf{\phantom{}3.8}$\pm$0.6\phantom{}\phantom{}\\
variational inference (B-PINN) & \phantom{}\textbf{0.111}$\pm$0.017\phantom{}\phantom{} & \phantom{}-31.4$\pm$11.5\phantom{-}\phantom{} & \phantom{}\textbf{0.005}$\pm$0.003\phantom{}\phantom{} & \phantom{}-392.6$\pm$409.4\phantom{-}\phantom{}\\
\bottomrule
\end{tabular}} 
\caption{
Evaluation metrics when solving the advection equation \eqref{eq:adv} with repulsion in ($f, \lambda$)-space, averaged over 5 runs with 25 ensemble members.
The mean values plus-or-minus one standard deviation are given.
The results for one of these runs are depicted in Figure \ref{fig:ds102_comp}.
\\
} 
\vspace{-1cm}
\label{tab:ds102_repf}
\end{table}
\end{DIFnomarkup}

Another drawback of VI is that the width of the uncertainty intervals remains arbitrary, since the weighting factor of the KL term between prior and posterior in the VI loss function \eqref{eq:loss_VI} needs to be hand-tuned.
The reason for this is that the prior put on the network parameters has no obvious relation to function space, and should therefore not be given too much weight.
In total, we conclude that VI does not yield truly Bayesian posteriors.

\begin{figure}[t]
    \centering
    \includegraphics[width=1\linewidth]{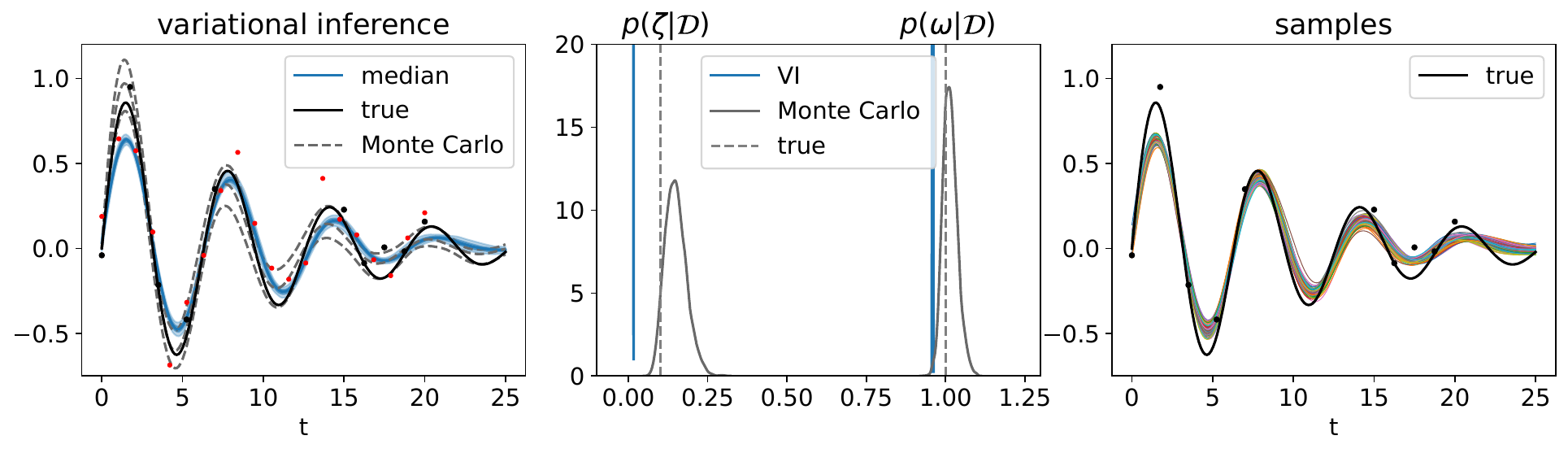}
    \captionof{figure}{ 
    Solving the damped harmonic oscillator \eqref{eq:dHO} with variational inference.
    \textbf{Left:}
    The predictions in $f$-space are depicted.
    The light- and dark-shaded areas give the regions between the [0.1, 0.9]- and the [0.25, 0.75]-quantiles, respectively.
    The dashed lines depict the 0.1- and 0.9-quantiles as obtained via MCMC, together with the median.
    Black dots correspond to train data and red dots to test data.
    \textbf{Middle:}
    The predictions in $\lambda$-space are depicted.
    The VI distribution is compared to the distribution obtained via MCMC.
    \textbf{Right:}
    Samples from the VI posterior are depicted.
    }
    \label{fig:ds5_MC_comp_VI}
\end{figure}

\section{Conclusions} \label{sec:conclusions}

In this work, we introduced the RE-PINN, and investigated the application of repulsive ensembles for uncertainty estimation to PINNs.
Building on the work of \citet{d2021repulsive}, we derived an expression for the repulsive term in PINNs and showed that DE parameters need to be included in the repulsion.
The repulsion results in an ensemble distribution that accurately represents the Bayesian posterior in the limit of many ensemble members.
While this holds in theory, different approximations are necessary to make the approach tractable in practice.
We investigated different approximations of the repulsive term and found that factorizing the density into lower-dimensional components improves the performance of the repulsive ensemble.

We performed experiments on a range of basic, low-dimensional DEs.
The experiments demonstrate that the standard ensemble vastly underestimates the uncertainty in PINN predictions.
The reason for this presumably lies in the fact that the PINN loss acts as an additional regularizer that further discourages diversity in the ensemble members, beyond the fact that the ensemble members converge to MAP estimates.
Including repulsion is therefore essential in the PINN context and significantly improves the results.
In accordance with the theoretical derivation from Section \ref{sec:method}, applying repulsion in the joint space of function values $f$ and DE parameters $\lambda$ works best.
The fully factorized model gave the most consistently good results, which we therefore recommend for usage.
The model predictions agree well with the Monte Carlo baselines, where available, confirming that repulsive ensembles can be considered Bayesian.

Our approach also outperformed B-PINN, the VI approach introduced by \citet{yang2021bpinn}, by a significant margin.
The latter is lacking in sample diversity, yields arbitrary widths of the uncertainty intervals, and often exhibits a degradation in the sample quality as compared to the MAP estimate.
The RE-PINN, on the other hand, is well equipped to combine both sample diversity and sample accuracy.
Since each of the ensemble members has its own set of parameters, distinct solutions can be learned accurately.
In the RE-PINN, the weighting of the repulsive term is also chosen in a more principled manner, leading to uncertainty estimates in agreement with the Bayesian posterior.
These properties make repulsive ensembles a natural choice for uncertainty estimation in PINNs.

While our experiments establish RE-PINN as a promising method for uncertainty quantification in PINNs, it is important to note the limitations of our investigation.
We focused on simple DEs, which has the advantage that an MC baseline for the Bayesian posterior could be obtained in many cases, allowing for a detailed evaluation of the accuracy of the posterior estimates.
These experiments show that RE-PINN outperforms existing methods for uncertainty quantification in PINNs, which already fail for these examples, but it is still unclear how the method will perform in case of more complex DEs or higher-dimensional problems.
A deeper examination of this constitutes a compelling direction for future work.

In the future, it would also be interesting to further explore possible ways of approximating the ensemble density.
Using Gaussian processes to approximate the ensemble density $\rho(f)$ could take correlations between values in $f$-space into account effectively and may reduce the number of ensemble members necessary for obtaining good results.
Extending the approach to infer the noise strength, or the noise distribution as in \citet{pilar2024physics}, together with the solution and DE parameters constitutes another avenue for future research.
The approach of \citet{zou2025learning} could presumably also be improved with repulsion, as it would encourage the differently initialized PINNs to converge to different DE solutions.
It would also be interesting to experiment with different methods of repulsion, such as repulsion in the space of input-gradients \citep{trinh2024inputgradient}, instead of the function values.
Investigating ways to enable sequential training of the ensemble members constitutes another promising direction of research.
It might be possible to store particle trajectories of preceding runs in a low-dimensional space, e.g., in ($f, \lambda$)-space, and use these for repulsion when training the remaining ensemble members.

\section*{Acknowledgments}
This work was undertaken at Uppsala University, Aalto University, Chalmers University of Technology and University of Gothenburg.
It is financially supported by the Swedish Research Council (VR) via the project Physics informed machine learning (registration number: 2021-04321).

\bibliographystyle{unsrtnat}
\bibliography{main}

\appendix
\newpage

\include{appendices}

\end{document}

%% file: appendices.tex
\section{Background}

Additional details on methods and results utilized in the main paper are given in this appendix.

\subsection{Derivation of the repulsive term} \label{app:derivation}
In this section, we outline the arguments that were employed in Section 3.2 of \citet{d2021repulsive} to derive the repulsive term in \eqref{eq:particle_dynamics_f}.

They consider the space of probability measures $\mathcal{P}_2(\mathcal{M})$ on the manifold $\mathcal{M}$,
\begin{equation}
    \mathcal{P}_2{\mathcal{M}} = \left\{ \varphi: \mathcal{M} \rightarrow [0,\infty) \middle| \int_{\mathcal{M}} d\varphi = 1, \int_{\mathcal{M}} \lvert x \rvert^2 \varphi(x) dx < +\infty \right\},
\end{equation}
together with the Wasserstein metric
\begin{equation}
    W_2^2(\mu, \nu) = \underset{\pi \in \Pi(\mu, \nu)}{\text{inf}} \int |x-y|^2 d\pi(x,y).
\end{equation}
They then proceed to employ the KL divergence to match the ensemble density $\rho(x)$ to the target distribution $\pi(x)$:
\begin{equation}
    \underset{\pi \in \mathcal{P}_2{\mathcal{M}}}{\text{inf}} D_{KL}(\rho, \pi) = \int_{\mathcal{M}}  \big(\log \rho(x) - \log \pi(x)\big)\rho(x) dx.
\end{equation}
When minimizing the KL divergence, the evolution of the measure $\rho$ (also referred to as the Wasserstein gradient flow), is given by the Liouville equation, which in turn simplifies to the Fokker-Planck equation:
\begin{align}
    \frac{\partial \rho(x)}{\partial t} &= \nabla \cdot  \Big(\rho(x) \nabla \big( \log \rho(x) - \log \pi(x)\big)\Big) \\
    &= - \nabla \cdot  \left(\rho(x) \nabla \log \pi(x) \right) + \Delta \rho(x).
\end{align}
This equation admits $\pi(x)$ as the unique stationary solution.
The corresponding deterministic particle dynamics ODE, for samples $x_i \sim \rho$, is then given by
\begin{equation}
    \frac{dx_i}{dt} = - \nabla \big(\log \rho(x_i) - \log \pi(x_i)\big),
\end{equation}
or, after discretizing it,
\begin{equation}
    x_i^{t+1} = x_i^t + \epsilon_t \left( \nabla \log \pi(x_i^t) - \nabla \log \rho(x_i^t) \right).
\end{equation}

In the context of repulsive ensembles, $\rho$ denotes the ensemble distribution and $\pi(x)$ is chosen as the posterior $p(x|\mathcal{D})$.
Note that, during the derivation, no restrictions have been placed on the domain of $\rho(x)$.
Hence, $x$ can be chosen as the network parameters $\theta$, or the function values $\f$, depending on which quantity is of interest.
This also vindicates us in writing down \eqref{eq:update_discrete_0} when considering $\rho(\f, \lambda)$ instead of $\rho(\f)$.

\section{Kernel density estimation} \label{app:KDE}
Given samples $x_i$, the density $\rho(x)$ can be approximated using kernel density estimation (KDE).
We then obtain the following expression:
\begin{equation}
    \rho(x) \approx \frac{1}{\sqrt{|h|} N_e} \sum_{i=1}^{N_e} k\left( \frac{x-x_i}{h} \right),
\end{equation}
where $h$ is a vector containing the bandwidth for each dimension of $x$;
the division in the argument of the kernel is elementwise.
The bandwidths are estimated with the median heuristic \citep{liu2016stein}, $h_d=\frac{\text{med } r^d_{ij}}{\sqrt{\log N_e}}$, where $r^d_{ij}$ denotes the pairwise distances between samples along the given dimension $d$.
We use the RBF kernel, $k(z) = \frac{1}{\sqrt{2 \pi}}e^{-\frac{\lVert z \rVert^2}{2}}$.
The case of joint densities $\rho(y, z)$ can be reduced to the given case via $x = \begin{bmatrix}
    y \\ z
\end{bmatrix}$.

\section{Monte Carlo estimates of the posterior} \label{app:MC}

The true posterior, $p(f, \lambda|\mD) = \frac{p(\mD|f, \lambda)p(f, \lambda)}{p(\mD)}$, is in general intractable.
In the case of ODEs, however, the solution is uniquely determined by the initial conditions $\{f_{0j}\}_{j=0}^{d}$ and the ODE parameters $\{\lambda_j\}_{j=0}^{N_{\lambda}}$.
Hence, it is possible to conduct Markov Chain Monte Carlo (MCMC) in the resulting $(d+N_{\lambda})$-dimensional space to obtain samples from $p(f_0, \lambda|\mathcal{D})$.
MCMC is considered the gold standard for Bayesian inference 
\citep{brooks2011handbook, nemeth2021stochastic} and produces samples from the true posterior.
Therefore, the thus obtained estimate can be considered a proxy for the ground truth PINN posterior.

The reason why the MCMC estimate can only be considered a proxy lies in the fact that the MC algorithm acts on exact ODE solutions, whereas PINNs include the ODE as a soft constraint.
Nevertheless, as PINNs also aim to obtain DE solutions that are as accurate as possible, the MC estimate would be exact for a perfectly tuned PINN.
In our experiments, we have weighted the DE constraints strongly towards the end of the training to ensure that deviations from exact DE solutions are only minor, and that the MC posterior remains a valid ground truth solution for the Bayesian posterior.

For example, in the case of the exponential equation \eqref{eq:exp}, the current state of the chain would be defined by $(f_0, \lambda)$.
The corresponding solution $f(t) = f_0 e^{\lambda t}$ can be employed to calculate the likelihood $p(\mD|f_0, \lambda)$.
Together with the prior $p(f_0, \lambda) = p(f_0)p(\lambda)$, all of the quantities required for evaluating a proposal are available.

For the MCMC algorithm, we made the following choices of priors:
for the exponential equation from Section~\ref{sec:exp}, we chose $p(f_0) \sim \mathcal{U}(-10,10)$ and $p(\lambda) \sim \mathcal{U}(-10,10)$.
For the damped harmonic oscillator from Section~\ref{sec:dHO}, we chose $p(f_0) \sim \mathcal{U}(0, 1.5)$, $p(\omega) \sim \mathcal{U}(0, 3)$, and $p(\zeta) \sim \mathcal{U}(0,0.9)$.
When training the PINNs, the same priors can be used by plugging them into~\eqref{eq:loss}.
However, to ensure stable training with useful gradients, it is necessary to use smoothened versions of the priors, i.e., to use Sigmoids instead of step functions.

We used the pyro package \citep{bingham2018pyro} for the implementation of the Monte Carlo algorithm, and chose Hamiltonian Monte Carlo (HMC) \citep{neal2011mcmc} with the No-U-Turn Sampler (NUTS) \citep{hoffman2014no}. 

\section{Variational inference} \label{app:VI}

We conduct VI as described in \citet{yang2021bpinn}, as a variant of the B-PINN.
Here, the network parameters are not fixed weights, but instead distributions to be learned.
We employ the commonly used mean-field Gaussian approximation, where the variational posterior density is factorized into independent Gaussian distributions: $q(\bt; \zeta) = \prod_j q_j(\bt_j; \zeta_{\mu, j}, \zeta_{\sigma, j})$, where $\bt = \{ \theta, \lambda\}$ contains both, the network parameters $\theta$ and the DE parameters $\lambda$.
The means $\zeta_{\mu}$ and standard deviations $\ln(1 + \exp (\zeta_{\sigma}))$ of these Gaussians are the parameters to be optimized during training.
The loss function for BNN-VI is given by
\begin{align}
    \mathcal{L}(\zeta) &= \frac{1}{N_z} \sum_{i=1}^{N_z} \ln q(\bt^{(i)}) - \ln p(\bt^{(i)}) - \ln p(\mathcal{D}|\bt^{(i)}), \\
    &\approx D_{\rm KL} (q(\bt), p(\bt)) + \frac{1}{N_z} \sum_{i=1}^{N_z} \mathcal{L}_f(\bt^{(i)}) + \mathcal{L}_{\mathcal{F}}(\bt^{(i)}), 
\end{align}
where the $\bt^{(i)}$ denote $N_z$ sets of parameters sampled from the variational posterior $q$.
Hence, the term $\frac{1}{N_z} \sum_i \ln q(\bt^{(i)}) - \ln p(\bt^{(i)}) \approx  D_{\rm KL} (q(\bt), p(\bt))$ gives an estimate of the KL divergence between the variational posterior and the prior.
When assuming Gaussian priors, it can be evaluated analytically.
The likelihoods $\mathcal{L}_f$ and $\mathcal{L}_{\mathcal{F}}$ are given in \eqref{eq:PINN_losses}.
After rescaling, the loss can be written in terms of the same weighting factors as in \eqref{eq:loss}:
\begin{equation} \label{eq:loss_VI}
    \mathcal{L}(\zeta) = \frac{1}{N_d N_z} \sum_{i=1}^{N_z} \sum_{j=1}^{N_d}  \left(\hat f_i(x_j) - y_j \right)^2 + \frac{f_{\lambda}}{N_c N_z} \sum_{i=1}^{N_z} \sum_{j=1}^{N_c}  \left(\mathcal{F}( \hat f_i(x_j), \lambda_i)\right)^2 + f_{\rm KL} D_{\rm KL} (q(\bt), p(\bt)).
\end{equation}
The factor $f_{\rm KL}$ constitutes an ad hoc weighting factor for the KL term, which was required for obtaining reasonable results;
omitting it led to standard deviations in the NN parameters that were too large and which led to nonsensical predictions that did not fit the data nor the DE well.

The performance of VI could likely be improved by choosing a more sophisticated posterior distribution.
However, determining a suitable choice would introduce significantly more tuning work than in the RE-PINN.
Another advantage of the RE-PINN over VI is that inference can be conducted directly in function space.
For VI, priors need to be defined in weight space, and it is not obvious what a good choice would be.
In our experiments, we chose zero-mean Gaussian priors with standard deviation $0.1$ for all of the network parameters.


\begin{table} [!t]
\resizebox{\textwidth}{!}{%
\begin{tabular}{l c c c c}
\toprule
Model & exponential equation & damped HO & advection & Lotka-Volterra \\ 
\midrule
non-repulsive & \textbf{0.073}$\pm$0.023 & \textbf{0.011}$\pm$0.002 & 0.055$\pm$0.007 & 0.330$\pm$0.021 \\
repulsive (f) & 0.135$\pm$0.017 & 0.013$\pm$0.002 & 0.055$\pm$0.007 & 0.359$\pm$0.061 \\
repulsive ($\lambda$) & 0.089$\pm$0.019 & 0.014$\pm$0.002 & 0.062$\pm$0.006 & \textbf{0.313}$\pm$0.035 \\
repulsive (f and $\lambda$) & 0.116$\pm$0.016 & 0.111$\pm$0.002 & 0.055$\pm$0.008 & 0.335$\pm$0.067 \\
repulsive (f and $\lambda$) (factorized) & 0.123$\pm$0.017 & 0.013$\pm$0.002 & 0.062$\pm$0.006 & 0.359$\pm$0.054 \\
repulsive (f and $\lambda$) (fully factorized) & 0.080$\pm$0.021 & 0.012$\pm$0.002 & \textbf{0.054}$\pm$0.007 & 0.399$\pm$0.076 \\
B-PINN/BNN-VI & 0.220$\pm$0.026 & 0.037$\pm$0.004 & 0.077$\pm$0.016 & 0.816$\pm$0.081 \\
\bottomrule
\end{tabular}}
\captionof{table}{
Adherence of individual posterior samples to the DE, as measured by the mean DE residuals $\frac{1}{N_z} \sum_{i=1}^{N_z} 2 \sigma_{\mathcal{F}}^2 \mathcal{L}_{\mathcal{F}}^i$.
Repulsion does not significantly impact the quality of individual samples.
For samples from the VI posterior, the DE fulfillment is adversely affected.
}
\label{tab:de_residuals}
\end{table}

\section{Experiments}

In this appendix, we give background information about the experiments conducted in Section \ref{sec:experiments} and present an additional experiment.

\subsection{Training details} \label{app:training}

Fully-connected neural networks with tanh activation function have been employed for the individual ensemble members.
To better isolate the impact of repulsion on the training outcome, the ensemble members were initialized with the same values for the different models given in Table \ref{tab:repulsive_models}.
The Adam optimizer \citep{kingma2014adam} was used for all of the examples.
When training PINNs, it is often advantageous to start with a lower weight for the DE constraint and only increase it later in the training.
Similarly, only activating the repulsion later in the training, when the individual ensemble members have come close to convergence, can lead to improved KDE estimates and stabler training of the ensemble.

For the exponential equation (\ref{sec:exp}), the networks had 2 layers of width 20 and the training lasted $10\,000$ iterations with learning rate $0.01$.
The strength of the measurement noise was $\sigma_f = 2$.
The weighting factor $f_{\lambda}$ in \eqref{eq:loss} was chosen as 5 and then increased to 10 and 25 at iterations $7\,500$ and $9\,000$, respectively.
For the damped harmonic oscillator (\ref{sec:dHO}), 3 layers of width 20 were chosen and the training lasted $15\,000$ iterations with learning rate $0.01$.
The strength of the measurement noise was $\sigma_f = 1$.
The weighting factor $f_{\lambda}$ in \eqref{eq:loss} was chosen as 1 and then increased to 5 and 10 at iterations $5\,000$ and $7\,500$, respectively.
For this example, the repulsive term was only activated after $3\,000$ iterations.
For the advection equation (\ref{sec:adv}), 4 layers of width 100 were chosen and the training lasted $5\,000$ iterations with learning rate $0.003$, with weight decay set to $5 \times 10^{-4}$.
The strength of the measurement noise was $\sigma_f = 0.1$.
The weighting factor $f_{\lambda}$ in \eqref{eq:loss} was chosen as 0.1.
For the Lotka-Volterra equation (\ref{app:lv}), 3 layers of width 20 were chosen and the training lasted $15\,000$ iterations with learning rate $0.01$.
The strength of the measurement noise was $\sigma_f = 1$.
The weighting factor $f_{\lambda}$ in \eqref{eq:loss} was chosen as 0.5 and then increased to 1 and 2 at iterations $5\,000$ and $7\,500$, respectively.

When training BNN-VI (Appendix \ref{app:VI}), the same network architectures were employed.
For the exponential equation \eqref{eq:exp}, the training lasted $10\,000$ iterations with learning rate $0.01$.
The weighting factor $f_{\lambda}$ was initialized as $1$ and then increased to $2$ and $10$ after $1\,000$ and $5\,000$ iterations, respectively. 
The KL term was activated after $5000$ iterations with a weighting factor of $f_{\rm KL}=0.01$.
For the damped harmonic oscillator \eqref{eq:dHO}, the training lasted $10\,000$ iterations with learning rate $0.01$.
The weighting factor $f_{\lambda}$ was initialized as $1$ and then increased to $2$ and $10$ after $2000$ and $3\,000$ iterations, respectively. 
The KL term was activated after $3000$ iterations with a weighting factor of $f_{\rm KL}=0.00002$.
For the advection equation \eqref{eq:adv}, the training lasted $15\,000$ iterations with learning rate $0.003$.
The weighting factor $f_{\lambda}$ was initialized as $0.01$ and then increased to $0.1$ after $10\,000$ iterations. 
The KL term was activated after $10000$ iterations with a weighting factor of $f_{\rm KL}=10^{-7}$.
For the Lotka-Volterra equation \eqref{eq:lv}, the training lasted $10\,000$ iterations with learning rate $0.01$.
The weighting factor $f_{\lambda}$ was initialized as $0.1$ and then increased to $1$ and $2$ after $2\,000$ and $5\,000$ iterations, respectively. 
The KL term was activated after $5\,000$ iterations with a weighting factor of $f_{\rm KL}=0.00001$.

\subsection{Computational complexity and resources} \label{app:computational}

Training an ensemble of PINNs is significantly more computationally expensive than training a single network.
For a standard, non-repulsive ensemble, it is straightforward to see that the overall computational cost will increase by a factor of $N_e$.
The contribution of the additional loss term in repulsive ensembles requires further consideration.
In our analysis, we will focus on the fully factorized variant defined in Table~\ref{tab:repulsive_models}.
The other variants are expected to be of similar cost, as KDE in n-dimensional space exhibits the same $\mathcal{O}(n)$ scaling as performing KDE n times in one dimension, when considering simple kernel functions such as the RBF kernel.


In the fully factorized model, $(N_f + N_\lambda)$ additional terms are added to the loss function, each of which involves a KDE with $N_e$ terms.
The difference in computational complexity between a single member of the non-repulsive and the repulsive ensemble is therefore of $\mathcal{O}(N_e (N_f + N_{\lambda}))$.
The overall contribution of the repulsive term to the computational cost of the full ensemble will hence scale as $\mathcal{O}(N_e^2 (N_f + N_{\lambda}))$.
While this term will become dominant for large ensemble sizes, the standard PINN loss is expensive due to the DE loss term \eqref{eq:Ll} and is expected to remain dominant for small to intermediate values of $N_e$.
This hypothesis is corroborated by the results in Table \ref{tab:runtime}.

The experiments in this paper were conducted on a computer with NVIDIA GeForce Titan XP 12GB GPU and Intel(R) Core(TM) i7-6850K CPU @ 3.60GHz CPU.
The runtimes for the different experiments are given in Table~\ref{tab:runtime}.
Adding repulsion does not increase the runtime by a large amount.
The different repulsive models have very similar runtimes and are therefore not listed individually.

\begin{table*}[h]
\centering
\caption{Runtime (in hours) for one run contributing to the entries in Tables \ref{tab:ds1_Nx10_repf}-\ref{tab:lv}, where $N_e=25$.}
\label{tab:runtime}
\begin{tabular}{rcc}
\toprule
& non-repulsive & repulsive \\
\cmidrule{2-3}
exponential equation (\ref{sec:exp}) & 0.35  & 0.43 \\
damped harmonic oscillator (\ref{sec:dHO})  & 0.69 & 0.78 \\
advection equation (\ref{sec:adv}) & 0.39 & 0.46 \\
Lotka-Volterra (\ref{app:lv}) & 0.69 & 0.78 \\
\bottomrule
\end{tabular}
\end{table*}

\subsection{Exponential equation} \label{app:exp}

\begin{figure}
    \centering
    \includegraphics[width=\linewidth]{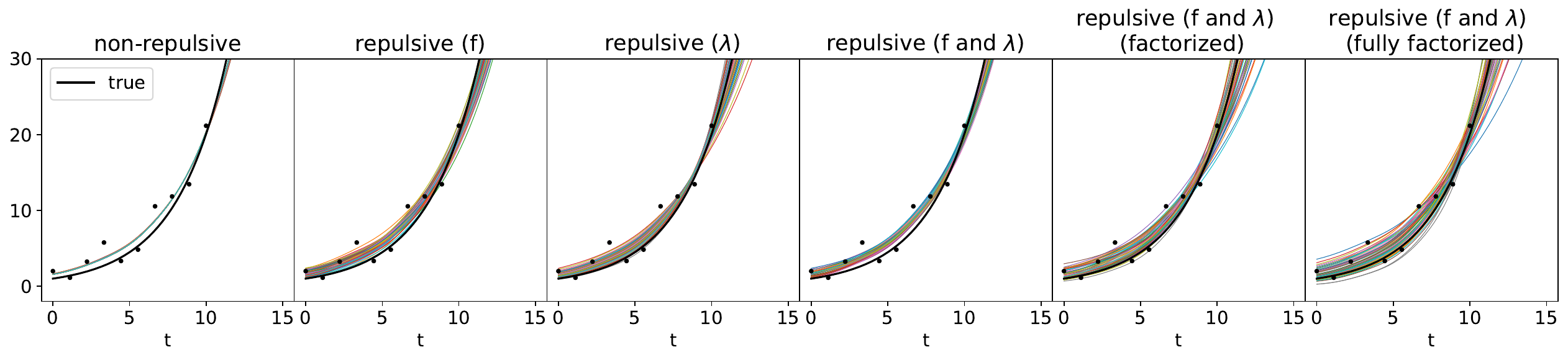}
    \caption{The ensemble members corresponding to the distributions in Figure \ref{fig:ds1_MC_comp} are depicted.}
    \label{fig:ds1_samples}
\end{figure}

In Figure \ref{fig:ds1_samples}, the individual ensemble members corresponding to Figure \ref{fig:ds1_MC_comp} are depicted.
In the plots, it can be observed that these curves adhere to the ODE individually, i.e. solutions that start at a higher initial value learned a lower value for the corresponding exponent $\lambda$ in the solution of \eqref{eq:exp} and eventually cross through other samples, which start at lower initial values but with larger $\lambda$.
This behavior also leads to the narrowing of the uncertainty intervals around $t=10$, observed in Figure~\ref{fig:ds1_MC_comp}.

\subsection{Damped harmonic oscillator} \label{app:dHO}

\begin{figure} [t]
    \centering
    \includegraphics[width=\linewidth]{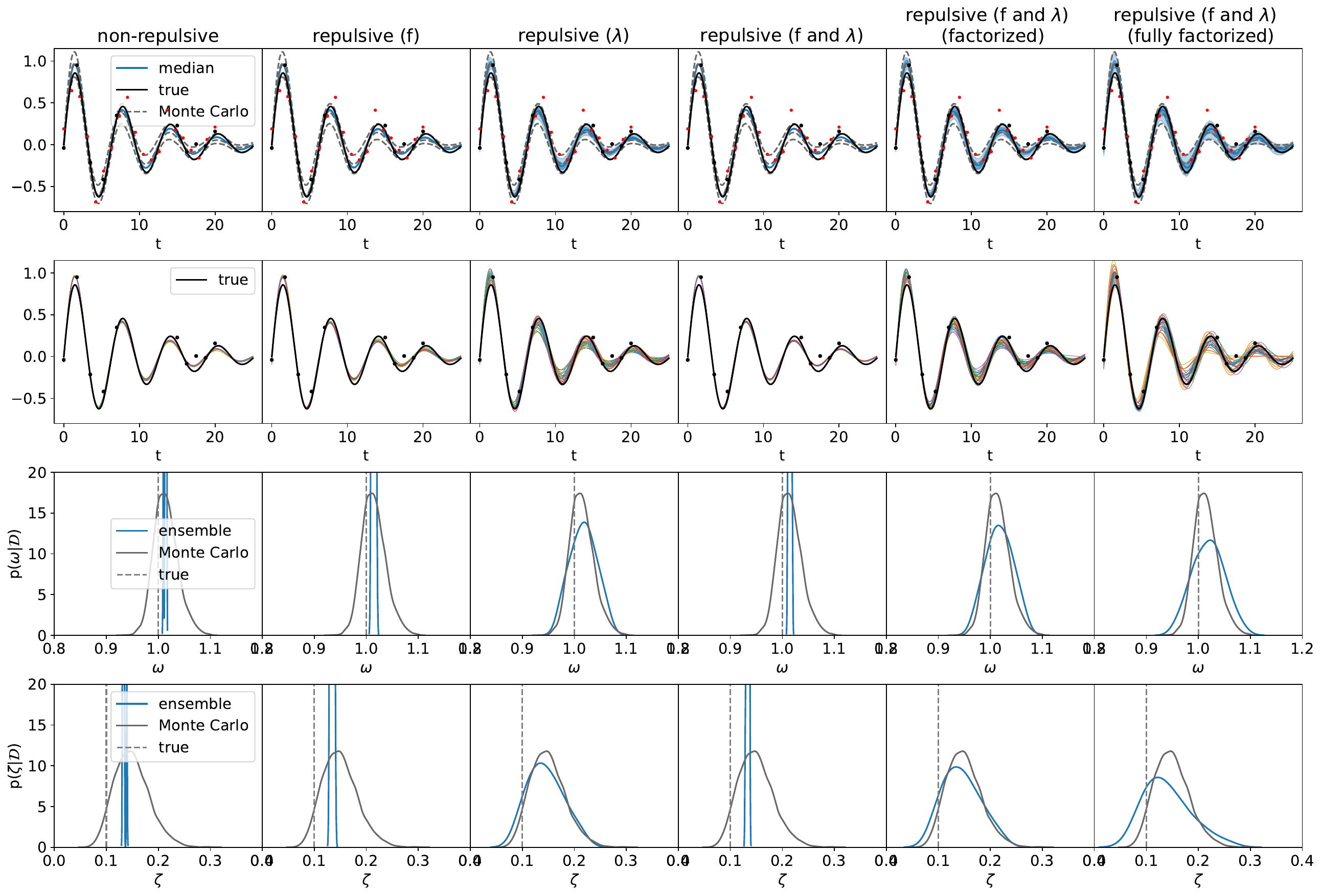}
    \caption{Solving the damped harmonic oscillator \eqref{eq:dHO} with repulsion in ($f$, $\lambda$)-space.
    \textbf{Top row:}
    The predictions in $f$-space are depicted.
    The light- and dark-shaded areas give the regions between the [0.1, 0.9]- and the [0.25, 0.75]-quantiles, respectively.
    The dashed lines depict the 0.1- and 0.9-quantiles as obtained via MCMC, together with the median.
    Black dots correspond to train data and red dots to test data.
    \textbf{Bottom row:}
    The predictions in $\lambda$-space are depicted.
    The ensemble distribution is compared to the distribution obtained via MCMC.
    The corresponding evaluation metrics are given in Table \ref{tab:dHO}.
    The ensembles have 25 members.}
    \label{fig:ds5_MC_comp_gap}
\end{figure}

Figure \ref{fig:ds5_MC_comp_gap} shows results for the damped harmonic oscillator corresponding to Figure \ref{fig:ds5_MC_comp} and Table \ref{tab:dHO} for all of the models.



\subsection{Advection equation} \label{app:adv}

\begin{figure}
    \centering
    \includegraphics[width=0.95\linewidth]{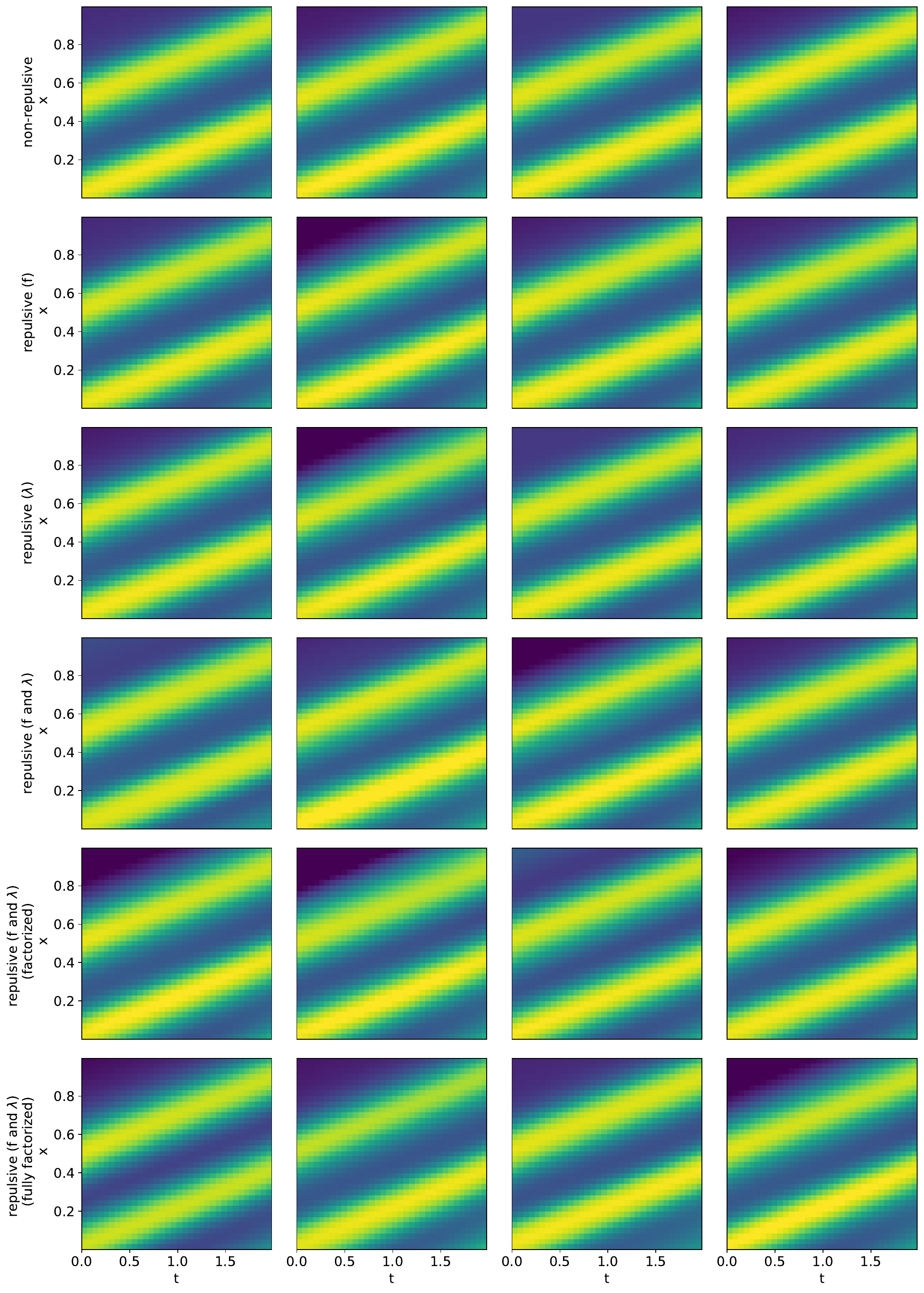}
    \caption{A subset of the ensemble members corresponding to the distributions in Figure \ref{fig:ds102_comp} is depicted.}
    \label{fig:ds102_samples}
\end{figure}

In Figure \ref{fig:ds102_samples}, individual members of the ensembles corresponding to the results in Section \ref{sec:adv} are depicted.
Also here, the different ensemble members constitute valid PDE solutions independently.

\subsection{The Lotka-Volterra equations} \label{app:lv}

The Lotka-Volterra equations describe predator-prey dynamics and are given by
\begin{subequations}\label{eq:lv}
\begin{align} 
    \frac{d x}{d t}  &= \alpha x - \beta x y, \\
    \frac{d y}{d t} &= -\gamma y + \delta x y.
\end{align}
\end{subequations}

The results are given in Figures \ref{fig:ds7_MC_comp} and Table \ref{tab:lv}.
The non-repulsive ensemble members again collapse to very similar solutions.
The different repulsive ensembles exhibit higher variation in their predictions and show altogether very similar performance.
They improve over the non-repulsive ensemble in terms of logL/N (test), and they yield good values on logL $\lambda$ more consistently.
MCMC did not converge for this example.

\begin{table} [b]
\resizebox{\textwidth}{!}{%
    \centering
    \includegraphics[width=1\linewidth]{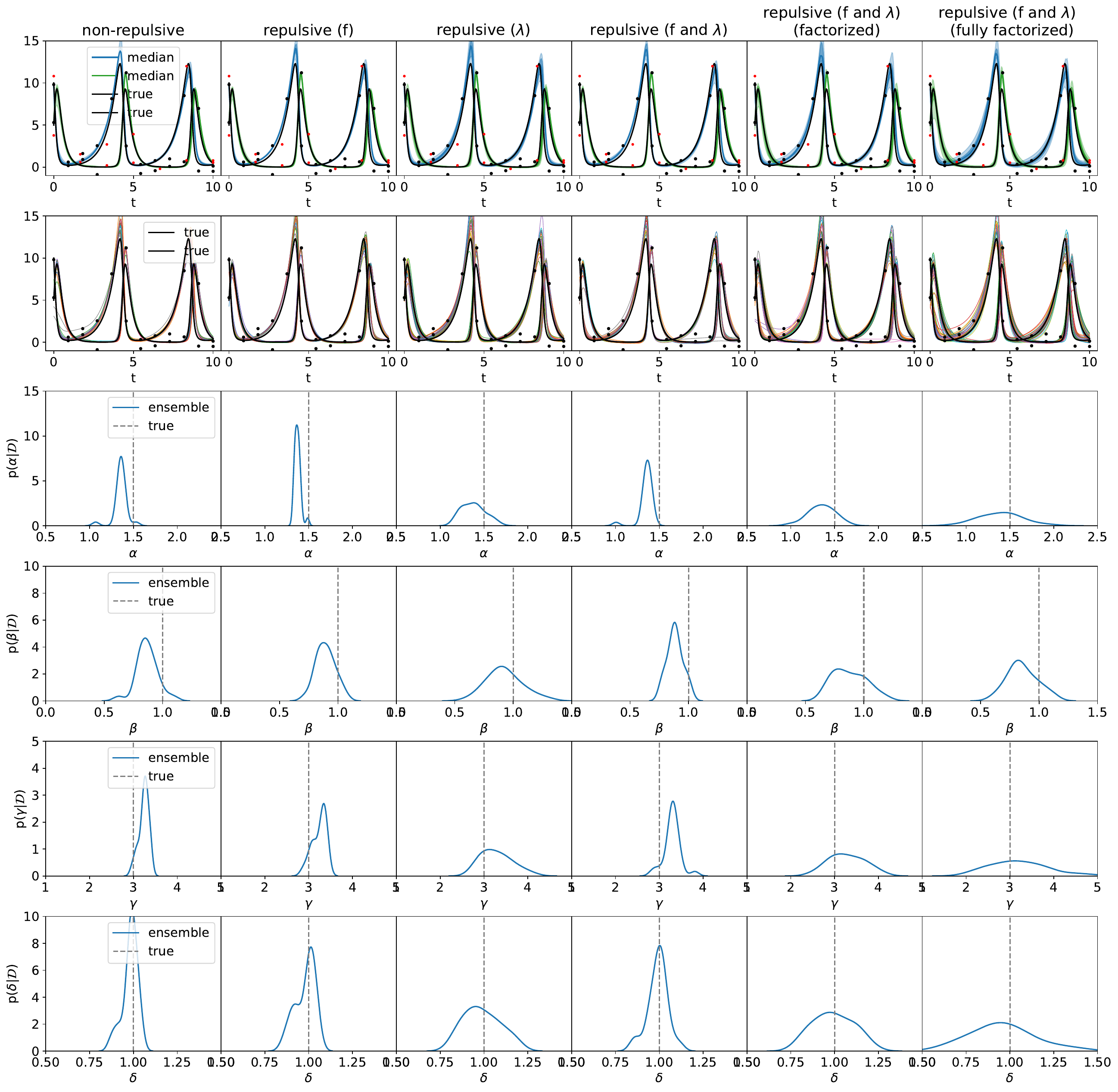}}
    \captionof{figure}{ 
    Solving the Lotka-Volterra equations \eqref{eq:exp} with repulsion in ($f$, $\lambda$)-space.
    \textbf{Top row:}
    The predictions in $f$-space are depicted.
    The light- and dark-shaded areas give the regions between the [0.1, 0.9]- and the [0.25, 0.75]-quantiles, respectively.
    Black dots correspond to train data and red dots to test data.
    \textbf{Bottom row:}
    The predictions in $\lambda$-space are depicted.
    The ensembles have 25 members.
    }
    \label{fig:ds7_MC_comp}
\resizebox{\textwidth}{!}{%
\begin{tabular}{l c c c c }
\hline
& RMSE (true) & logL/N (test) & $|\lambda - \hat \lambda|$ & logL $\lambda$ \\ 
\cline{2-5}
non-repulsive & \phantom{}0.54$\pm$0.10\phantom{}\phantom{} & \phantom{}-720.1$\pm$380.5\phantom{-}\phantom{} & \phantom{}0.13$\pm$0.04\phantom{}\phantom{} & \phantom{}0.98$\pm$2.05\phantom{}\phantom{}\\
repulsive (f) & \phantom{}0.55$\pm$0.11\phantom{}\phantom{} & \phantom{}-673.1$\pm$575.9\phantom{-}\phantom{} & \phantom{}0.13$\pm$0.03\phantom{}\phantom{} & \phantom{}-1.88$\pm$5.28\phantom{-}\phantom{}\\
repulsive ($\lambda$) & \textbf{\phantom{}0.52}$\pm$0.11\phantom{}\phantom{} & \phantom{}-267.0$\pm$151.8\phantom{-}\phantom{} & \phantom{}0.11$\pm$0.04\phantom{}\phantom{} & \textbf{\phantom{}1.92}$\pm$0.60\phantom{}\phantom{}\\
repulsive (f and $\lambda$) & \phantom{}0.54$\pm$0.10\phantom{}\phantom{} & \phantom{}-825.3$\pm$633.6\phantom{-}\phantom{} & \phantom{}0.12$\pm$0.03\phantom{}\phantom{} & \phantom{}-1.95$\pm$4.73\phantom{-}\phantom{}\\
repulsive (f and $\lambda$) (factorized) & \phantom{}0.56$\pm$0.07\phantom{}\phantom{} & \phantom{}-228.1$\pm$133.9\phantom{-}\phantom{} & \textbf{\phantom{}0.11}$\pm$0.04\phantom{}\phantom{} & \phantom{}1.81$\pm$0.43\phantom{}\phantom{}\\
repulsive (f and $\lambda$) (fully factorized) & \phantom{}0.54$\pm$0.09\phantom{}\phantom{} & \textbf{\phantom{}-39.4}$\pm$12.2\phantom{-}\phantom{} & \phantom{}0.12$\pm$0.02\phantom{}\phantom{} & \phantom{}0.15$\pm$0.54\phantom{}\phantom{}\\
B-PINN/BNN-VI & \phantom{}1.96$\pm$0.19\phantom{}\phantom{} & \phantom{}-575.4$\pm$193.0\phantom{-}\phantom{} & \phantom{}0.36$\pm$0.07\phantom{}\phantom{} & \phantom{}-7476.72$\pm$2000.14\phantom{-}\phantom{}\\
\hline
\end{tabular}}
\captionof{table}{
Evaluation metrics when solving Lotka-Volterra equation \eqref{eq:lv} with repulsion in ($f, \lambda$)-space, averaged over 5 runs with 25 ensemble members.
The mean values plus-or-minus one standard deviation are given.
The results for one of these runs are depicted in Figure \ref{fig:ds7_MC_comp}.
\\
}
\label{tab:lv}
\end{table}

%% file: main.bib
@article{d2021repulsive,
  title={Repulsive deep ensembles are {B}ayesian},
  author={D'Angelo, Francesco and Fortuin, Vincent},
  journal={Advances in Neural Information Processing Systems},
  volume={34},
  pages={3451--3465},
  year={2021}
}

@article{lakshminarayanan2017simple,
  title={Simple and scalable predictive uncertainty estimation using deep ensembles},
  author={Lakshminarayanan, Balaji and Pritzel, Alexander and Blundell, Charles},
  journal={Advances in neural information processing systems},
  volume={30},
  year={2017}
}

@article{jiang2023practical,
  title={Practical uncertainty quantification for space-dependent inverse heat conduction problem via ensemble physics-informed neural networks},
  author={Jiang, Xinchao and Wang, Xin and Wen, Ziming and Li, Enying and Wang, Hu},
  journal={International Communications in Heat and Mass Transfer},
  volume={147},
  pages={106940},
  year={2023},
  publisher={Elsevier}
}

@article{yang2022multi,
  title={Multi-output physics-informed neural networks for forward and inverse {PDE} problems with uncertainties},
  author={Yang, Mingyuan and Foster, John T},
  journal={Computer Methods in Applied Mechanics and Engineering},
  volume={402},
  pages={115041},
  year={2022},
  publisher={Elsevier}
}

@article{li2024bayesian,
  title={{B}ayesian neural network and {B}ayesian physics-informed neural network via variational inference for seismic petrophysical inversion},
  author={Li, Peng and Grana, Dario and Liu, Mingliang},
  journal={Geophysics},
  volume={89},
  number={6},
  pages={1--46},
  year={2024},
  publisher={Society of Exploration Geophysicists}
}

@inproceedings{tan2025evidential,
  author    = {Tan, Hai Siong and Wang, Kuancheng and McBeth, Rafe},
  title     = {Evidential Physics-Informed Neural Networks},
  booktitle = {Proceedings of the International Conference on Scientific Computing and Machine Learning (SCML 2025)},
  pages      = {22--31},
  year       = {2025}
}

@inproceedings{haitsiukevich2023improved,
  title={Improved training of physics-informed neural networks with model ensembles},
  author={Haitsiukevich, Katsiaryna and Ilin, Alexander},
  booktitle={2023 International Joint Conference on Neural Networks (IJCNN)},
  pages={1--8},
  year={2023},
  organization={IEEE}
}

@article{rover2024pinnferring,
  title={{PINN}ferring the {H}ubble function with uncertainties},
  author={R{\"o}ver, Lennart and Sch{\"a}fer, Bj{\"o}rn Malte and Plehn, Tilman},
  journal={energy},
  volume={5},
  pages={9},
  year={2024}
}

@article{raissi2019physics,
  title={Physics-informed neural networks: {A} deep learning framework for solving forward and inverse problems involving nonlinear partial differential equations},
  author={Raissi, Maziar and Perdikaris, Paris and Karniadakis, George E},
  journal={Journal of Computational physics},
  volume={378},
  pages={686--707},
  year={2019},
  publisher={Elsevier}
}

@article{cuomo2022scientific,
  title={Scientific machine learning through physics--informed neural networks: {W}here we are and what’s next},
  author={Cuomo, Salvatore and Di Cola, Vincenzo Schiano and Giampaolo, Fabio and Rozza, Gianluigi and Raissi, Maziar and Piccialli, Francesco},
  journal={Journal of Scientific Computing},
  volume={92},
  number={3},
  pages={88},
  year={2022},
  publisher={Springer}
}

@article{cai2021fluid,
  title={Physics-informed neural networks ({PINN}s) for fluid mechanics: {A} review},
  author={Cai, Shengze and Mao, Zhiping and Wang, Zhicheng and Yin, Minglang and Karniadakis, George Em},
  journal={Acta Mechanica Sinica},
  volume={37},
  number={12},
  pages={1727--1738},
  year={2021},
  publisher={Springer}
}

@article{lawal2022physics,
  title={Physics-informed neural network ({PINN}) evolution and beyond: {A} systematic literature review and bibliometric analysis},
  author={Lawal, Zaharaddeen Karami and Yassin, Hayati and Lai, Daphne Teck Ching and Che Idris, Azam},
  journal={Big Data and Cognitive Computing},
  volume={6},
  number={4},
  pages={140},
  year={2022},
  publisher={MDPI}
}

@article{cai2021heat,
  title={Physics-informed neural networks for heat transfer problems},
  author={Cai, Shengze and Wang, Zhicheng and Wang, Sifan and Perdikaris, Paris and Karniadakis, George Em},
  journal={Journal of Heat Transfer},
  volume={143},
  number={6},
  pages={060801},
  year={2021},
  publisher={American Society of Mechanical Engineers}
}

@article{takamoto2022pdebench,
  title={Pdebench: {A}n extensive benchmark for scientific machine learning},
  author={Takamoto, Makoto and Praditia, Timothy and Leiteritz, Raphael and MacKinlay, Daniel and Alesiani, Francesco and Pfl{\"u}ger, Dirk and Niepert, Mathias},
  journal={Advances in Neural Information Processing Systems},
  volume={35},
  pages={1596--1611},
  year={2022}
}

@article{zou2025learning,
  title={Learning and discovering multiple solutions using physics-informed neural networks with random initialization and deep ensemble},
  author={Zou, Zongren and Wang, Zhicheng and Em Karniadakis, George},
  journal={Proceedings of the Royal Society A: Mathematical, Physical and Engineering Sciences},
  volume={481},
  number={2325},
  year={2025},
  publisher={The Royal Society}
}

@article{yang2021bpinn,
  title={B-{PINN}s: {B}ayesian physics-informed neural networks for forward and inverse {PDE} problems with noisy data},
  author={Yang, Liu and Meng, Xuhui and Karniadakis, George Em},
  journal={Journal of Computational Physics},
  volume={425},
  pages={109913},
  year={2021},
  publisher={Elsevier}
}

@article{sahli2020physics,
  title={Physics-informed neural networks for cardiac activation mapping},
  author={Sahli Costabal, Francisco and Yang, Yibo and Perdikaris, Paris and Hurtado, Daniel E and Kuhl, Ellen},
  journal={Frontiers in Physics},
  volume={8},
  pages={42},
  year={2020},
  publisher={Frontiers Media SA}
}

@inproceedings{gal2016dropout,
  title={Dropout as a {B}ayesian approximation: {R}epresenting model uncertainty in deep learning},
  author={Gal, Yarin and Ghahramani, Zoubin},
  booktitle={international conference on machine learning},
  pages={1050--1059},
  year={2016},
  organization={PMLR}
}

@article{graves2011practical,
  title={Practical variational inference for neural networks},
  author={Graves, Alex},
  journal={Advances in neural information processing systems},
  volume={24},
  year={2011}
}

@book{box2011bayesian,
  title={Bayesian inference in statistical analysis},
  author={Box, George EP and Tiao, George C},
  year={2011},
  publisher={John Wiley \& Sons}
}

@inproceedings{roeder2021linear,
  title={On linear identifiability of learned representations},
  author={Roeder, Geoffrey and Metz, Luke and Kingma, Durk},
  booktitle={International Conference on Machine Learning},
  pages={9030--9039},
  year={2021},
  organization={PMLR}
}

@inproceedings{
wang2019function,
title={{F}unction {S}pace {P}article {O}ptimization for {B}ayesian {N}eural {N}etworks},
author={Ziyu Wang and Tongzheng Ren and Jun Zhu and Bo Zhang},
booktitle={International Conference on Learning Representations},
year={2019}
}

@article{jospin2022hands,
  title={Hands-on {B}ayesian neural networks--{A} tutorial for deep learning users},
  author={Jospin, Laurent Valentin and Laga, Hamid and Boussaid, Farid and Buntine, Wray and Bennamoun, Mohammed},
  journal={IEEE Computational Intelligence Magazine},
  volume={17},
  number={2},
  pages={29--48},
  year={2022},
  publisher={IEEE}
}

@article{jordan1998variational,
  title={The variational formulation of the {F}okker--{P}lanck equation},
  author={Jordan, Richard and Kinderlehrer, David and Otto, Felix},
  journal={SIAM journal on mathematical analysis},
  volume={29},
  number={1},
  pages={1--17},
  year={1998},
  publisher={SIAM}
}

@article{fang2023ensemble,
  title={Ensemble learning for physics informed neural networks: A gradient boosting approach},
  author={Fang, Zhiwei and Wang, Sifan and Perdikaris, Paris},
  journal={arXiv preprint arXiv:2302.13143},
  year={2023}
}

@article{kingma2014adam,
  title={Adam: {A} method for stochastic optimization},
  author={Kingma, Diederik P and Ba, Jimmy},
  journal={arXiv preprint arXiv:1412.6980},
  year={2014}
}

@inproceedings{trinh2024inputgradient,
title={Input-gradient space particle inference for neural network ensembles},
author={Trung Trinh and Markus Heinonen and Luigi Acerbi and Samuel Kaski},
booktitle={The Twelfth International Conference on Learning Representations},
year={2024}
}

@inproceedings{pilar2024physics,
  title={Physics-informed neural networks with unknown measurement noise},
  author={Pilar, Philipp and Wahlstr{\"o}m, Niklas},
  booktitle={6th Annual Learning for Dynamics \& Control Conference},
  pages={235--247},
  year={2024},
  organization={PMLR}
}

@book{neal2012bayesian,
  title={Bayesian learning for neural networks},
  author={Neal, Radford M},
  volume={118},
  year={2012},
  publisher={Springer Science \& Business Media}
}

@article{neal2011mcmc,
  title={{MCMC} using {H}amiltonian dynamics},
  author={Neal, Radford M and others},
  journal={Handbook of markov chain monte carlo},
  volume={2},
  number={11},
  pages={2},
  year={2011},
  publisher={Chapman and Hall/CRC}
}

@article{hoffman2014no,
  title={The {No-U-Turn} sampler: adaptively setting path lengths in {H}amiltonian {M}onte {C}arlo.},
  author={Hoffman, Matthew D and Gelman, Andrew and others},
  journal={J. Mach. Learn. Res.},
  volume={15},
  number={1},
  pages={1593--1623},
  year={2014}
}

@article{bingham2018pyro,
author = {Bingham, Eli and Chen, Jonathan P. and Jankowiak, Martin and Obermeyer, Fritz and
          Pradhan, Neeraj and Karaletsos, Theofanis and Singh, Rohit and Szerlip, Paul and
          Horsfall, Paul and Goodman, Noah D.},
title = {{Pyro: Deep Universal Probabilistic Programming}},
journal = {Journal of Machine Learning Research},
year = {2018}
}

@article{liu2016stein,
  title={Stein variational gradient descent: {A} general purpose {B}ayesian inference algorithm},
  author={Liu, Qiang and Wang, Dilin},
  journal={Advances in neural information processing systems},
  volume={29},
  year={2016}
}

@book{ambrosio2008gradient,
  title={Gradient flows: in metric spaces and in the space of probability measures},
  author={Ambrosio, Luigi and Gigli, Nicola and Savar{\'e}, Giuseppe},
  year={2008},
  publisher={Springer Science \& Business Media}
}

@article{lagaris1998artificial,
  title={Artificial neural networks for solving ordinary and partial differential equations},
  author={Lagaris, Isaac E and Likas, Aristidis and Fotiadis, Dimitrios I},
  journal={IEEE transactions on neural networks},
  volume={9},
  number={5},
  pages={987--1000},
  year={1998},
  publisher={IEEE}
}

@article{nemeth2021stochastic,
  title={Stochastic gradient {M}arkov {C}hain {M}onte {C}arlo},
  author={Nemeth, Christopher and Fearnhead, Paul},
  journal={Journal of the American Statistical Association},
  volume={116},
  number={533},
  pages={433--450},
  year={2021},
  publisher={Taylor \& Francis}
}

@book{brooks2011handbook,
  title={Handbook of{M}arkov {C}hain {M}onte {C}arlo},
  author={Brooks, Steve and Gelman, Andrew and Jones, Galin and Meng, Xiao-Li},
  year={2011},
  publisher={CRC press}
}
